%% file: main.tex
\pdfoutput=1
\documentclass[sigconf]{acmart}

\usepackage{lipsum}
\usepackage{listings}
\usepackage{extarrows}
\usepackage{subfigure}
\usepackage{xspace}
\usepackage{graphicx}
\usepackage{makecell}
\usepackage{ifthen}
\usepackage{xifthen}
\usepackage{wrapfig}
\usepackage{amsthm}
\usepackage{stmaryrd}
\usepackage{threeparttable}
\usepackage{enumerate}
\usepackage{multirow}
\usepackage{booktabs}
\usepackage{titlesec}
\usepackage{mathtools, nccmath}
\usepackage{comment}
\usepackage{amsmath,amsfonts}
\usepackage[framemethod=TikZ]{mdframed}
\usepackage{lipsum}
\usepackage{float}
\usepackage{threeparttable}
\usepackage{caption}
\usepackage{tcolorbox}
\usepackage{listings}
\usepackage{upquote}
\usepackage[T1]{fontenc}
\usepackage{xcolor}
\usepackage[symbol]{footmisc}
\usepackage{svg}
\usepackage{enumerate}
\usepackage{enumitem}
\usepackage{wrapfig}

\newcommand{\model}{WebGLM\xspace}
\newcommand{\dataset}{WebGLM-QA\xspace}
\newcommand{\vpara}[1]{\vspace{0.04in}\noindent\textbf{#1}\xspace}
\newcommand{\hide}[1]{} 

\AtBeginDocument{%
  \providecommand\BibTeX{{%
    \normalfont B\kern-0.5em{\scshape i\kern-0.25em b}\kern-0.8em\TeX}}}

\copyrightyear{2023}
\acmYear{2023}
\setcopyright{acmlicensed}
\acmConference[KDD '23] {Proceedings of the 29th ACM SIGKDD Conference on Knowledge Discovery and Data Mining}{August 6--10, 2023}{Long Beach, CA, USA.}
\acmBooktitle{Proceedings of the 29th ACM SIGKDD Conference on Knowledge Discovery and Data Mining (KDD '23), August 6--10, 2023, Long Beach, CA, USA}
\acmPrice{15.00}
\acmISBN{979-8-4007-0103-0/23/08}
\acmDOI{10.1145/3580305.3599931}

\usepackage[ruled,vlined]{algorithm2e}

\begin{document}

\title{WebGLM: Towards An Efficient Web-Enhanced Question Answering System with Human Preferences}


\author{Xiao Liu}
\email{liuxiao21@mails.tsinghua.edu.cn}
\affiliation{
  \institution{Tsinghua University}
 \city{Beijing}
 \country{China}
}
\authornote{XL, HL, and HY contributed equally and this work was done when HY interned at Tsinghua. $^\dagger$Corresponding Authors: YD and JT.}

\author{Hanyu Lai}
\authornotemark[1]
\email{laihy19@mails.tsinghua.edu.cn}
\affiliation{
  \institution{Tsinghua University}
 \city{Beijing}
 \country{China}
}

\author{Hao Yu}
\authornotemark[1]
\email{yuhao2019@buaa.edu.cn}
\affiliation{
  \institution{Beihang University}
 \city{Beijing}
 \country{China}
}

\author{Yifan Xu}
\email{xuyifan2001@gmail.com}
\affiliation{
  \institution{Tsinghua University}
 \city{Beijing}
 \country{China}
}

\author{Aohan Zeng}
\email{zah22@mails.tsinghua.edu.cn}
\affiliation{
  \institution{Tsinghua University}
 \city{Beijing}
 \country{China}
}

\author{Zhengxiao Du}
\email{zx-du20@mails.tsinghua.edu.cn}
\affiliation{
  \institution{Tsinghua University}
 \city{Beijing}
 \country{China}
}

\author{Peng Zhang}
\email{peng.zhang@zhipuai.cn}
\affiliation{
  \institution{Zhipu.AI}
 \city{Beijing}
 \country{China}
}

\author{Yuxiao Dong}
\email{yuxiaod@tsinghua.edu.cn}
\affiliation{
  \institution{Tsinghua University}
 \city{Beijing}
 \country{China}
}
\authornotemark[2]

\author{Jie Tang}
\email{jietang@tsinghua.edu.cn}
\affiliation{
  \institution{Tsinghua University}
  \city{Beijing}
  \country{China}
}
\authornotemark[2]
\renewcommand{\shortauthors}{Liu and Lai and Yu, et al.}

\begin{abstract}
  We present WebGLM, a web-enhanced question-answering system based on the General Language Model (GLM). 
  Its goal is to augment a pre-trained large language model (LLM) with web search and retrieval capabilities while being efficient for real-world deployments. 
  To achieve this, we develop WebGLM with strategies for the LLM-augmented retriever, bootstrapped generator, and human preference-aware scorer. 
  Specifically, we identify and address the limitations of WebGPT (OpenAI), through which WebGLM is enabled with accuracy, efficiency, and cost-effectiveness advantages. 
  In addition, we propose systematic criteria for evaluating web-enhanced QA systems.
  We conduct multi-dimensional human evaluation and quantitative ablation studies, which suggest the outperformance of the proposed WebGLM designs over existing systems. 
  \model with the 10-billion-parameter GLM (10B) is shown to perform better than the similar-sized WebGPT (13B) and even comparably to WebGPT (175B) in human evaluation.
  The code, demo, and data are at \url{https://github.com/THUDM/WebGLM}.
\end{abstract}



\maketitle
    
\renewcommand{\thefootnote}{\arabic{footnote}}

\input{sections/1_introduction.tex}
\input{sections/2_related_works.tex}
\input{sections/3_methodology.tex}

\input{sections/6_metrics.tex}
\input{sections/8_evaluation.tex}
\input{sections/conclusion.tex}

\section*{ACKNOWLEDGEMENT}
This work is supported by Technology and Innovation Major Project of the Ministry of Science and Technology of China under Grant 2022ZD0118600 and 2022ZD0118601, NSF of China for Distinguished Young Scholars (No. 61825602), NSF of China (No. 62276148), and a research fund from Zhipu.AI.

\bibliographystyle{ACM-Reference-Format}
\bibliography{references}
\clearpage
\appendix

\input{sections/appendix.tex}



\end{document}

%% file: sections/1_introduction.tex
\begin{figure}[t!]
    \includegraphics[width=0.98\linewidth]{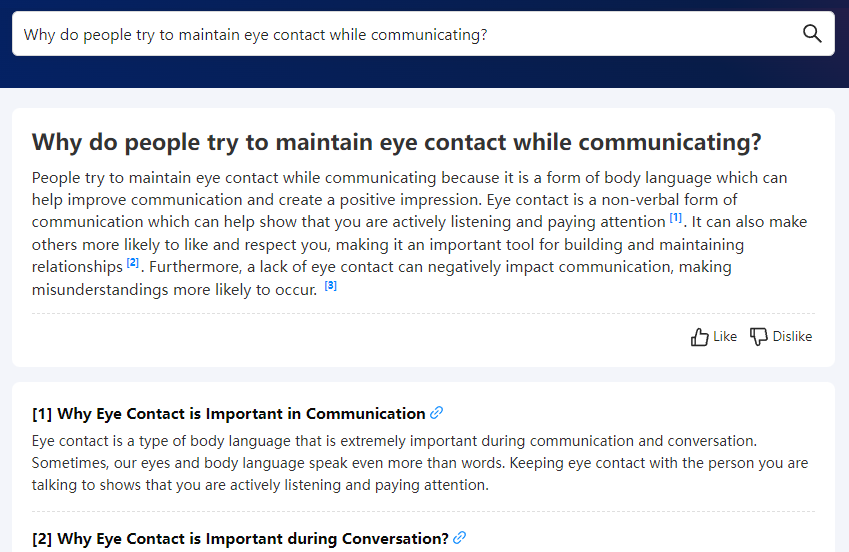}
    \caption{A screenshot of \model's response to an example question with web references.} 
    \vspace{-5mm}
    \label{fig:intro_example}
\end{figure}

\begin{figure}[t!]
    \includegraphics[width=0.92\linewidth]{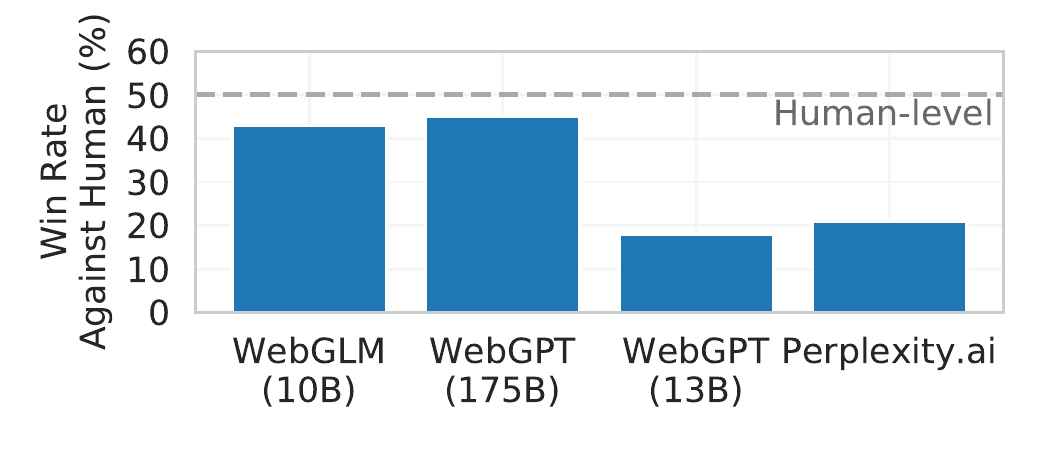}
    \vspace{-5mm}
    \caption{The win rates of popular web-enhanced QA systems against human references. \textmd{\model (10B) performs comparably to WebGPT (175B), approaching human-level QA ability.}} 
    \vspace{-5mm}
    \label{fig:intro_figure}
\end{figure}

\hide{
\begin{figure}[t!]
    \includegraphics[width=0.92\linewidth]{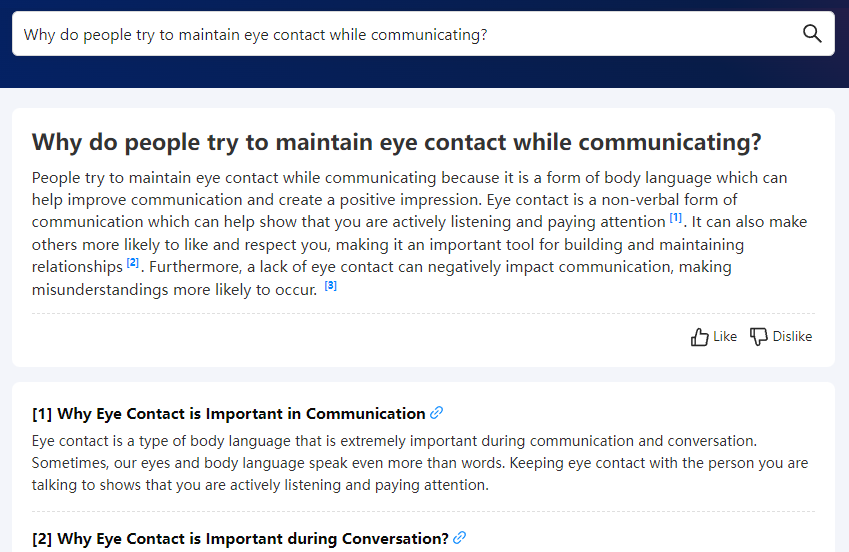}
    \includegraphics[width=0.97\linewidth]{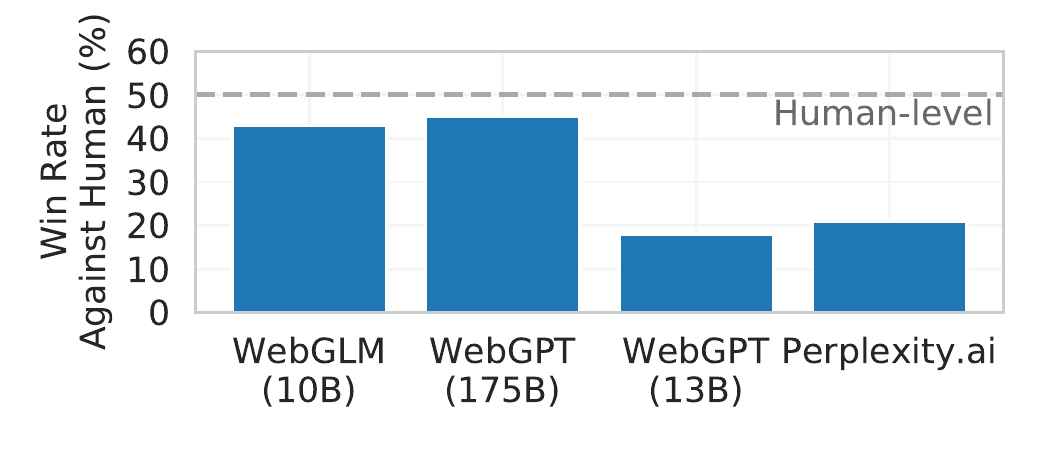}
    \vspace{-5mm}
    \caption{(Upper) A screenshot of \model's response to a certain question; (Bottom) The win rates of popular web-enhanced QA systems against human references. \textmd{\model (10B) performs comparably to WebGPT (175B), approaching human-level QA ability.}} 
    \vspace{-5mm}
    \label{fig:intro_figure}
\end{figure}

}

\section{Introduction}
Large language models (LLMs), such as GPT-3~\cite{gpt3}, PaLM~\cite{chowdhery2022palm}, OPT~\cite{opt}, BLOOM~\cite{bloom}, and GLM-130B~\cite{glm-130b}, have significantly pushed the boundary of machines' ability on language understanding and generation.
Question answering~\cite{squad,natural_questions}, one of the most fundamental language applications, has also been substantially advanced by the recent LLM developments.
Existing studies suggest that the performance of LLMs' closed-book QA~\cite{roberts2020much} and in-context learning QA~\cite{gpt3,liu-etal-2022-makes} is comparable to supervised models, furthering our understanding on LLMs' potential to memorize knowledge.

However, even for LLMs, their capacity is not unlimited, and when it comes to challenges that require sufficient rare-knowledge, LLMs fail to meet up human expectations.
Hence recent efforts have been focused on constructing LLMs augmented from external knowledge, such as retrieval~\cite{realm,rag,atlas} and web search~\cite{webgpt}.
For example, WebGPT~\cite{webgpt} can browse the web, answer complex questions in long form, and provide useful references correspondingly.

Despite its success, the original WebGPT method~\cite{webgpt} is far from real-world deployments.
First, it relies on abundant expert-level annotations of browsing trajectories, well-written answers, and answer preference labeling, requiring considerable expenses, time, and training.
Second, the behavior cloning method (i.e., imitation learning) requires its base model GPT-3 to emulate human experts by instructing the system to interact with a web browser, issue operation commands (e.g., \texttt{Search}, \texttt{Read}, and \texttt{Quote}), and then retrieve relevant information from online sources.
Finally, the multi-turn nature of web browsing demands intensive computation resources and can be too slow for user experience, e.g., costing about 31 seconds for WebGPT-13B to response a 500-token prompt. 

In this work, we present \model---a practical web-enhanced QA system based on the 10-billion-parameter General Language Model (GLM-10B)~\cite{du2022glm}. 
An example is illustrated in Figure \ref{fig:intro_example}. 
It is efficient, cost-effective, human preference-aware, and most importantly, of comparable quality to WebGPT.
The system employs multiple new strategies and designs to achieve good performance, including:

\vpara{An LLM-augmented Retriever}: a two-staged retriever that implements coarse-grained web search and fine-grained LLM-distilled retrieval.
It is inspired by the fact that LLMs like GPT-3 can naturally learn to adopt correct references, and such ability could be distilled to improve smaller dense retrievers.

\vpara{A Bootstrapped Generator}: a GLM-10B based answer generator that is trained on quoted long-formed QA samples and bootstrapped by LLM in-context learning. 
We discover that instead of relying on expensive human expert writing in WebGPT, LLMs can be enabled to learn to generate high-quality data with proper citation-based filtering. 

\vpara{A Human Preference-aware Scorer}: a scorer, that is trained over online QA forums' user thumb-up signals, is able to learn human majority preferences on different answers. 
Compared to WebGPT's expert labeling, we prove that a proper dataset construction could also produce a high-quality scorer.

Our extensive human evaluation and quantitative ablation results demonstrate the efficiency and effectiveness of the \model system. 
Specifically, WebGLM (10B) surpasses the similar-scaled WebGPT (13B) and performs comparably to WebGPT (175B) on our Turing test (Cf. Figure~\ref{fig:intro_figure}). 
\model's improvement against the only publicly-available system---Perplexity.ai---also makes it among the best public web-enhanced QA systems as of this submission.

To sum up, in this paper, we make the following contributions:

\begin{itemize}
    \item We construct \model, an efficient web-enhanced QA system with human preferences. 
    It significantly outperforms the similar-sized WebGPT (13B) and performs comparably to WebGPT (175B). 
    It also surpasses Perplexity.ai---a popular system powered by LLMs and search engines.
    \item We identify WebGPT's limitations on real-world deployments. 
    We propose a set of new designs and strategies to allow \model's high accuracy while achieving efficient and cost-effective advantages over baseline systems.
    \item We formulate the human evaluation metrics for evaluating web-enhanced QA systems. Extensive human evaluation and experiments demonstrate \model's strong capability and also generate insights into the system's future developments.
\end{itemize}

\hide{

\begin{figure}[t!]
    \includegraphics[width=0.92\linewidth]{figures/demo.png}
    \includegraphics[width=0.97\linewidth]{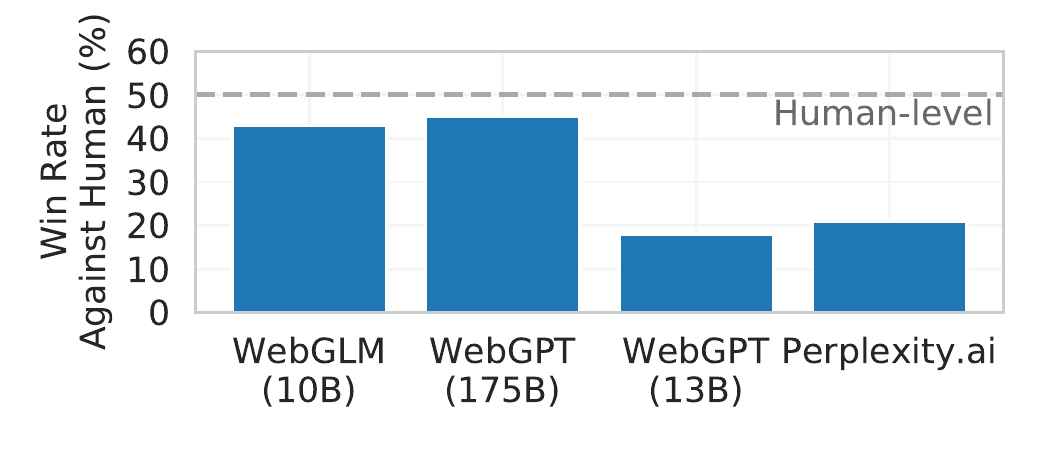}
    \vspace{-5mm}
    \caption{(Upper) A screenshot of \model's answering; (Bottom) Win rates of existing web-enhanced QA systems against human reference. \textmd{\model (10B) performs comparably to WebGPT (175B), approaching human-level QA ability.}} 
    \vspace{-5mm}
    \label{fig:intro_figure}
\end{figure}

\section{Introduction}
Large language models (LLMs), such as GPT-3~\cite{gpt3}, PaLM~\cite{chowdhery2022palm}, OPT~\cite{opt}, BLOOM~\cite{bloom}, and GLM-130B~\cite{glm-130b}, have significantly pushed the boundary of machine's ability on language understanding and generation.
Question answering~\cite{squad,natural_questions}, one of the most attractive and fundamental language applications, has also been substantially advanced by the LLM development.
Typical discoveries, including LLMs' closed-book QA~\cite{roberts2020much} and in-context learning QA performance~\cite{gpt3,liu-etal-2022-makes} that is comparable to supervised models, refresh people' understanding on their potential to memorize knowledge.

However, LLMs' capacity is limited, and when it comes to challenges that require sufficient rare knowledge, LLMs fail to meet up our expectations.
Hence a series of recent efforts in the community have been focused on constructing language models augmented from retrieval~\cite{realm,rag,atlas} and web search~\cite{webgpt}.
These models take advantage of external knowledge to accomplish challenging missions that are beyond previous imagination.
For example, WebGPT~\cite{webgpt} can browse the web, answer complex human questions in long form, and provide useful references correspondingly (e.g., Cf Figure~\ref{fig:intro_figure}).

Despite its success, the WebGPT is far from real deployment.
First, it relies on abundant expert-level annotations of browsing trajectories, well-written answers, and answer preference labeling, which requires considerable expenses, time, and training.
Second, the behavior cloning method (i.e., imitation learning) requires GPT-3 to emulate human experts by instructing the system to interact with a web browser, issue operation commands (e.g., \texttt{Search}, \texttt{Read}, and \texttt{Quote}), and then retrieve relevant information from online sources.
The browsing' multi-turn nature demands intensive computation resources and can be too slow for user experience.
Even with its answer quality, to construct an efficient web-enhanced QA system remains a huge challenge.

In this work, we present \model---a practical web-enhanced QA system based on GLM-10B---that is efficient, cost-effective, human preference-aware, and most importantly, of comparable quality to WebGPT (Cf. Figure~\ref{fig:intro_figure}).
The system employs multiple new strategies and designs to achieve the performance, including:

\vpara{LLM-augmented Retriever}: a two-staged retriever that implements coarse-grained web search and fine-grained LLM-distilled retrieval.
It is especially inspired by the fact that LLMs like GPT-3 can naturally learn to adopt correct references, and such ability could be distilled to improve smaller dense retrievers.

\vpara{Bootstrapped Generator}: a GLM-10B based answer generator that is trained on LLM in-context learning bootstrapped quoted long-formed QA samples, in which we discover that with proper citation-based filtering, LLMs can learn to generate high-quality data that originally depends on expensive human expert writing.

\vpara{Human Preference-aware Scorer}: a scorer that is trained over online QA forums' user thumb-ups to learn human majority preference on different answers. Compared to WebGPT's expert labeling, we prove that proper dataset construction could allow a high-quality scorer either.

Our extensive multi-dimensional human evaluation and quantitative ablation results demonstrate the efficiency and effectiveness of the \model system, which significantly surpasses similar-scaled WebGPT (13B) and performs comparably to WebGPT (175B) on our Turing test.
\model's improvement against the only publicly available system---Perlexity.ai\footnote{\url{https://perplexity.ai/}}---also makes it the best public web-enhanced QA system now.

To sum up, in this paper, we make the following contributions:

\begin{itemize}
    \item We construct \model, an efficient web-enhanced QA system with human preference. It significantly outperforms similar-sized WebGPT (13B) and performs comparably to WebGPT (175B). It also surpasses the only public similar system Perplexity.ai.
    \item We identify the limitations for WebGPT to be really deployed, and propose a set of new designs and strategies to allow \model's high performance while keeping the efficiency and cost-effectiveness.
    \item We set up the human evaluation metrics for evaluating web-enhanced QA systems. Extensive human evaluation and experiments demonstrate \model's capability and also provide deep insights to the system's future development.
\end{itemize}

}

%% file: sections/2_related_works.tex
\section{Related Work}
The construction of web-enhanced QA systems is a systematic project that requires cross-domain collaboration, including large language models, open-domain question answering, retrieval augmentation, and reinforcement learning from human feedback.
Here we briefly introduce related literature on them.

\vpara{Large Language Models (LLMs).}
Self-supervised~\cite{liu2021self} LLMs have attracted plenty of attention in nowadays natural language processing (NLP). 
Their huge number of parameters captures and stores versatile knowledge~\cite{liu2021gpt} and enables their outstanding performance on various challenges. 
Typical LLMs include GPT-3~\cite{gpt3}, PALM~\cite{chowdhery2022palm}, OPT~\cite{opt}, BLOOM~\cite{bloom}, and GLM-130B~\cite{glm-130b}. 
One of the fascinating LLM properties is prompt-based in-context learning (ICL), 
which allows tuning-free task transfer via prepended demonstration samples.
Recent works have been focusing on the optimization~\cite{liu-etal-2022-makes,su2022selective, Min2022NoisyCL, zhao2021calibrate} and analysis~\cite{min2022rethinking,Xie2021AnEO,rubin2022learning} of ICL.

\vpara{Open-domain Question Answering (Open QA).}
Traditional QA datasets such as SQuAD~\cite{squad} assume the reference is available. 
On the contrary, open-domain QA targets the open world and is more practical but challenging.
For example, Natural Questions~\cite{natural_questions} dataset consists of queries from the Google search engine and annotations from Wikipedia paragraphs.
Web Questions~\cite{berant2013semantic} derives open-domain questions from knowledge bases.
MS Marco~\cite{ms_marco} gathers passage texts and corresponding labels to questions.

However, most Open QA datasets and models are limited to answer short answer phrases, while people usually prefer more informative long-formed answers with references.
A possible reason is that constructing and evaluating long-formed QA datasets with open-world references are difficult, requiring expert-level annotations.
Recent attempts include ELI5~\cite{eli5} that collects queries and long-formed answers with scores from Reddit and WebGPT~\cite{webgpt} which hires groups of experts and leverages up to 175-billion-parameter GPT-3 as the backbone.
WebGLM aims to provide another effective and cost-effective solution for the challenge.
 
\vpara{Retrieval-augmentation.}
Mainstream information retrieval approaches include sparse-vector-based BM25 and TF-IDF, and the recent dense-vector-based methods such as DPR~\cite{dpr} and Contriever~\cite{izacard2022unsupervised}. 
The idea of retrieval-augmented language models introduced by REALM~\cite{realm} argues the joint optimization of retriever and language modeling. 
Following representative works include RAG~\cite{rag}, Fusion-in-Decoder~\cite{fid}, and Atlas~\cite{atlas}.
The idea of WebGPT also loosely falls into the field, as it asks the LLM to interact with the browser to seek relevant information for better accuracy.
Nevertheless, it can cost intensive computation and is too slow for practical deployment.
In this work, WebGLM tackles the problem efficiently by distilling LLMs' knowledge to smaller retrievers.

\vpara{Reinforcement Learning from Human Feedback (RLHF).}
Automated scoring of text generation is a well-established area of research. 
BLEU~\cite{bleu} and ROUGE~\cite{rouge} take into account the overlap ratio between the target and reference.
METEOR~\cite{meteor} considers the accuracy and recall rate of the whole corpus. 
Other methods, such as BERTScore~\cite{bertscore}, evaluate using cosine similarity of contextual embedding from deep language models. 
In recent years, some work advocates learning scorers from human feedback~\cite{learning-to-summarize,instructgpt} via asking models to predict human preference. 
The scorers, or namely reward models, can be used to optimize the text generator via reinforcement learning.
Such methods, which WebGPT is also affiliated with, have achieved great success in real-world applications.

%% file: sections/3_methodology.tex
\begin{figure*}[t]
\includegraphics[width=0.95\linewidth]{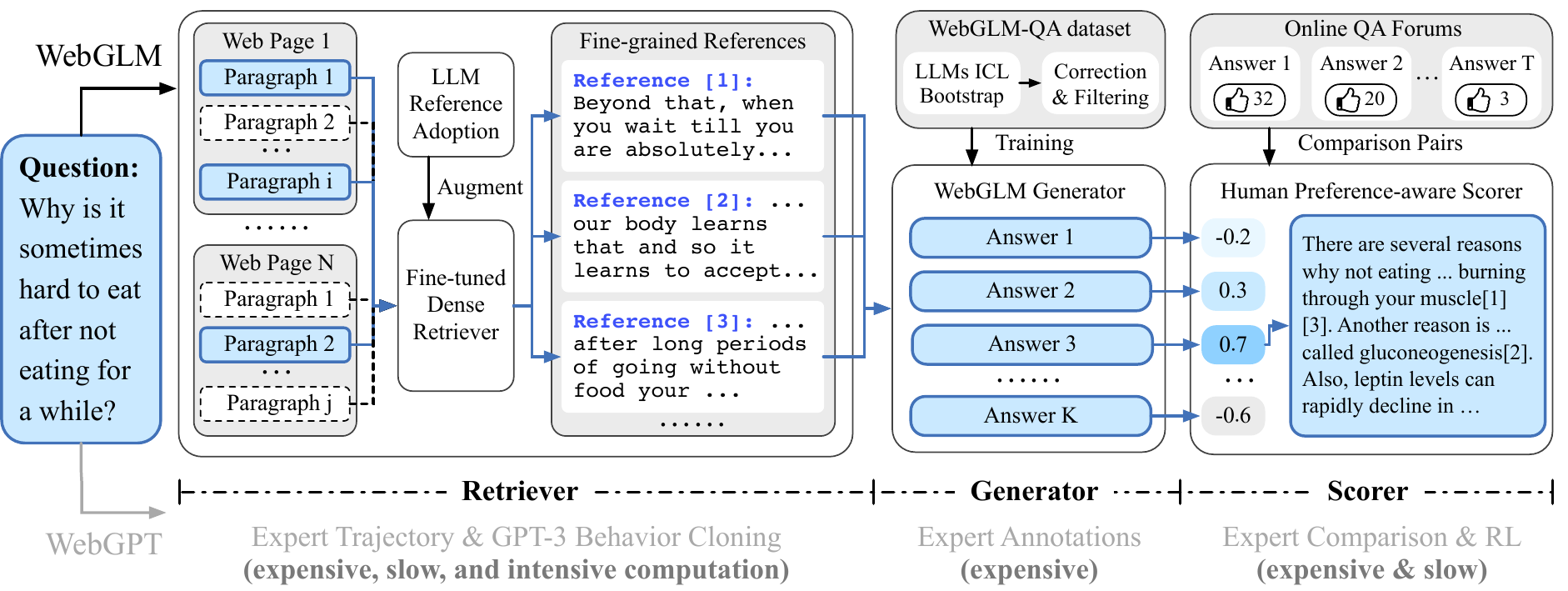}
\vspace{-2mm}
\caption{\model system pipeline. \textmd{Our system includes three sub-modules: LLM-augmented retriever recalls the top-5 most relevant paragraphs as the reference sources; Bootstrapped generator yields answers according to the question and reference sources; Human preference-aware scorer assesses all answers and picks the highest-scored one as the final result. Compared to WebGPT, \model is a more efficient and cost-effective web-enhanced QA system with comparable answer quality.}} \label{fig:main_process}
\vspace{-2mm}
\end{figure*}

\section{The WebGLM System}
Constructing an LLM-based web-enhanced QA system can be expensive and challenging. The
web information is rich but noisy for certain queries, and creating high-quality human answers with references for training can be outrageously expensive.
This type of systems usually involves three critical components: retriever, generator, and scorer.

Take WebGPT~\cite{webgpt} as an example, which employs experts for dataset annotation.
Its retriever leverages GPT-3 to ``behavior-clone'' human experts' web-browsing trajectory to search, read, and quote.
In addition, the generator is trained on expert-written long answers with references.
And finally, the scorer learns to predict experts' preferences over different answers, and its scores serve as rewards for the generator's reinforcement learning.
Despite WebGPT's primary success, its retrieval can be slow, and the data annotations required for training the generator and scorer are too costly, significantly hindering its wide public adoptions.

In this work, we aim to build an efficient web-enhanced QA system that understands human preferences for actual use.
To combine the advantages of LLMs and well-established open QA studies, we present a series of new designs and strategies for our web-enhanced QA system \model based on GLM~\cite{du2022glm}:

\begin{itemize}[leftmargin=*,itemsep=0pt,parsep=0.2em,topsep=0.2em,partopsep=0.0em]
    \item \textbf{An LLM-augmented Retriever}: we design two stages: coarse-grained web search and fine-grained LLM-augmented dense retrieval~\cite{izacard2022unsupervised}, for finding relevant references given queries.
    \item \textbf{A Bootstrapped Generator}: we derive \dataset, an LLM-bootstrapped quoted and long-formed QA dataset via in-context learning and corresponding strategies to clean and refine. It includes 45k high-quality after filtering and 83k noisy but diverse samples before filtering. The backbone of \model system is a GLM model trained on the dataset.
    \item \textbf{A Human Preference-aware Scorer}: we develop techniques to learn human majority preference from online QA forums' thumb-ups instead of expensive expert feedback, and successfully train a human preference-aware scorer for best-of-n selection.
\end{itemize}

The LLM API used for research purpose in this work is text-davinci-003 unless specified.
In the following sections, we will introduce the algorithm and implementation details of each component, which finally form the \model pipeline sequentially.

\begin{figure}[t]
    \includegraphics[width=0.85\linewidth]{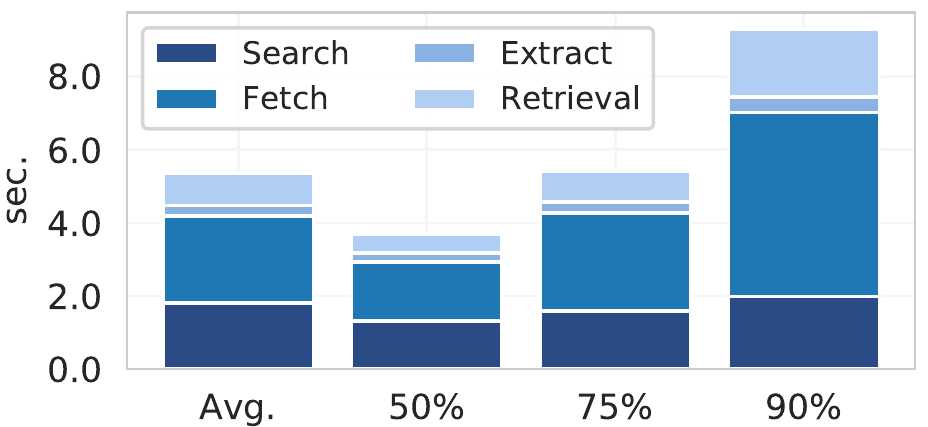}
    \vspace{-3mm}
    \caption{\model retriever time analysis. \textmd{50\% of queries can be done within 4.0s, and 90\% of them can be loaded within 10.0s. Most of time is spent on fetching web pages after searching.}} 
    \vspace{-5mm}
    \label{fig:search_time}
\end{figure}

\subsection{LLM-augmented Retriever}
In conventional open QA, the systems usually only retrieve from reliable sources (e.g., Wikipedia) and fail to benefit from whole web-scale knowledge.
However, the flip side of the coin is that wild web pages can be hard to acquire and purify.
In \model, we make attempts to solve the problem via two-stage retrieval: coarse-grained web search and fine-grained LLM-augmented retrieval.

\subsubsection{Coarse-grained Web Search}~\\
We leverage third-party web search engines (i.e., Google API) to acquire primary candidate web page URLs.
In most cases, from our observation, these pages can cover the necessary contexts and knowledge to answer questions besides considerably abundant irrelevant information.
The procedures are shown in Figure~\ref{fig:main_process}. 
Specifically, it can be roughly divided into three steps: 

\begin{enumerate}[leftmargin=*,itemsep=0pt,parsep=0.2em,topsep=0.2em,partopsep=0.0em]
    \item \textbf{Search}: At this stage, we enter the question into the search API and will obtain a list of URLs for potentially-relevant pages (usually less than 10).
    \item \textbf{Fetch}: Then, we crawl the corresponding HTML contents according to the URLs obtained. Since there are many candidate pages, we improve efficiency through parallel crawling.
    \item \textbf{Extract}: Next, based on HTML2TEXT\footnote{https://github.com/aaronsw/html2text}, we extract the part of text contents in the HTML pages and divide them into a list of paragraphs according to line breaks.
\end{enumerate}

Since the web crawl usually takes sufficient time, we have paid great efforts to optimize the speed of the component to allow user-acceptable responding speed (Cf. Figure~\ref{fig:search_time}).
For example, in the ``Fetch'' step, if the page is loaded synchronously, the loading time will be 2-3 minutes long. 
The parallel asynchronous enables the quick loading of most pages in 5s (about 98\%).

\subsubsection{Fine-grained LLM-augmented Retrieval}~\\
Through the first three stages, we have retrieved a number of potential contexts to questions. 
However, many of them are still irrelevant even under the filtering of widely-used dense retrievers (in our trial, up to 30\% of top-ranked contexts are unrelated).
As a solution, WebGPT~\cite{webgpt} uses behavior cloning (i.e., imitation learning) to leverage LLMs' strong language comprehensibility for reference selection.
Notwithstanding its effectiveness, the strategy is slow in deployment and expensive in labeling.

\vpara{LLMs' Reference Adoption.}
To mitigate the issue, we propose to combine smaller retrievers' efficiency and LLMs' strong ability to distinguish.
We take Contriever~\cite{izacard2022unsupervised} as the smaller retriever in \model, an unsupervised pre-trained model that encodes texts into embeddings and retrieves by finding the maximum inner product pair of them.
We transfer LLMs' natural property of reference adoption to small retrievers to improve them.

\begingroup
\input{tables/natural_adoption.tex}
Specifically, we find LLMs can naturally distinguish and only adopt useful references in in-context learning (ICL).
We create a 200-query dataset, where each query is accompanied with 5 top-ranked candidate references from Contriever.
We manually annotate the relevance of each piece of reference (Cf.  Table~\ref{tab:natural_adoption}). We find only 68.6\% of them are related.
However, when we provide the query with corresponding candidate references to GPT-3 for 1-shot in-context learning inference (see details in Section~\ref{sec:generator}), we discover that the LLM would only adopt part of the references and the corresponding accuracy is 90.2\%, far better than Contriever's.

\endgroup

\vpara{Augmentation Implementation.}
To transfer the reference adoption knowledge from GPT-3 to Contriever, we leverage the GPT-3's reference adoption from our bootstrapped dataset \dataset to additionally fine-tune Contrievers.
As the reference marks generated by GPT-3 can be wrong sometimes, we use the citation correction method based on Rouge-1 precision to match quotations and references (see those details in Section~\ref{sec:generator}).
Therefore, the labels we use for training are the Rouge-1 precision scores of a query-reference pair.

In the fine-tuning, we use two Contrievers to encode questions and references individually, and compute their inner products as the predictions.
We leverage Mean Square Error (MSE) as the loss function for the predictions and Rouge-1 precision scores to train the Contrievers.
Our further quantitative experiment demonstrates that the augmentation significantly improves Contriever web-enhanced QA retrieval accuracy (see Table~\ref{tab:contriever-2-performance} for details).

\subsubsection{Speed analysis}~\\
Retrieval is no doubt the most time-consuming part in any web-scale QA system.
A slow QA system, whatever high its accuracy is, would spoil the user experience.
We report the speed of each steps in our LLM-augmented retriever. 

We sample a subset from ELI5~\cite{eli5} test set to retrieve and calculate the average, the median, 75\% quantile, 90\% quantile, and 99\% quantile time spent in each step. 
From Figure~\ref{fig:search_time}, we can know that our average time spent is about 5.3s, the median total time spent is about 4.07s, and 90\% of searches can be loaded in 10s. 
The main bottleneck of our retrieval is in the second step of fetching each page, when we have to request multiple web pages from different sources.
Consequently, due the contents of various pages on the network are different, some pages take very long time to load, or just cannot be returned correctly.

In Appendix \ref{sec:efficiency}, we conduct a more detailed analysis of retrieval efficiency and point out that the retrieval efficiency of WebGLM is far better than that of WebGPT.

\begin{figure*}[t]
    \centering
    \includegraphics[width=\linewidth]{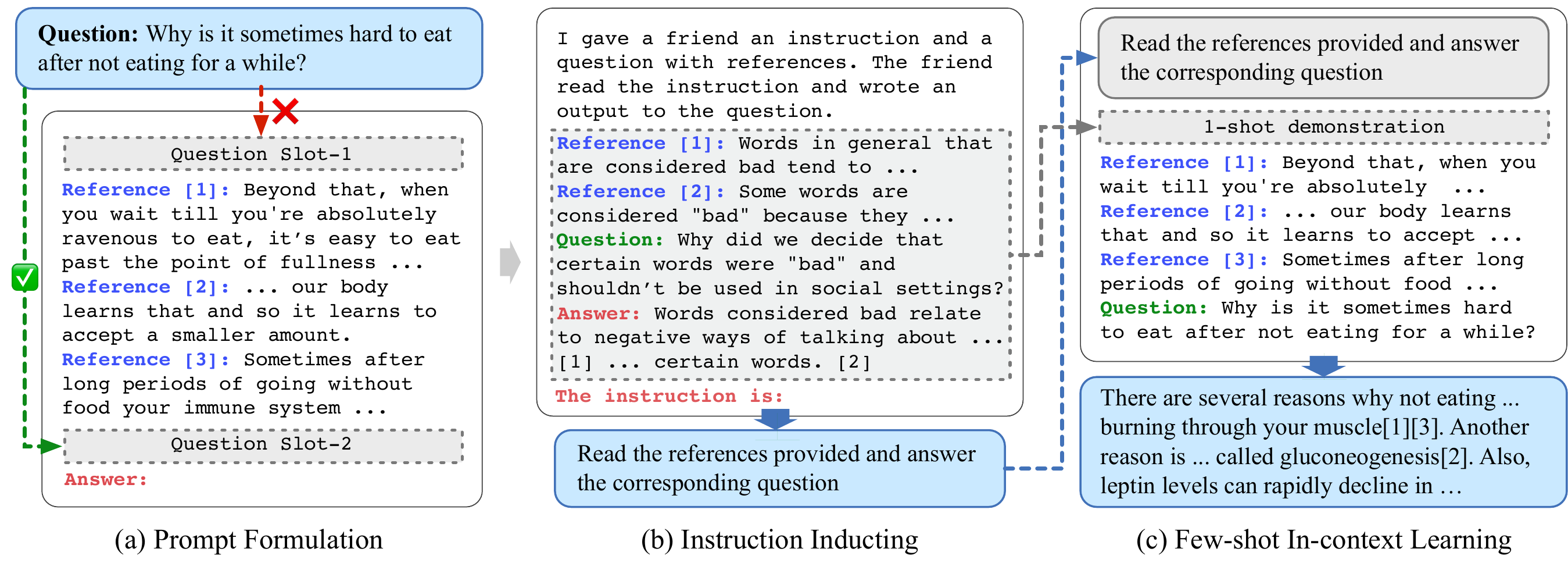}
    \vspace{-6mm}
    \caption{We construct \dataset for generator training via LLM in-context bootstrapping. \textmd{It includes three stages: 1) prompt formulation, 2) instruction inducting, and 3) few-shot in-context learning. In this way, we avoid the outrageous cost in time and money for hiring experts but still create a high-quality quoted long-formed QA dataset.}}
    \vspace{-3mm}
    \label{fig:bootstrapping}
\end{figure*}

\subsection{Bootstrapped Generator} \label{sec:generator}
A major obstacle in building web-enhanced QA system is the high cost for curating expert-level QA datasets that are long-formed and properly cited.
Compared to traditional or free-formed QA, we expect the system to yield fact-grounded answers with correct references (see example in~\ref{fig:bootstrapping}).
WebGPT reports to hire a group of full-time experts to write answers for training, which is far beyond ordinary budgets.

Fortunately, LLMs' in-context learning~\cite{gpt3,chowdhery2022palm}, which refers to their capabilities to transfer to new tasks conditioned on few in-context samples, have been demonstrated and well-explored recently.
Thus we propose to bootstrap large amounts of quoted long answers via leveraging a few high-quality answers, LLMs, questions from ELI5~\cite{eli5}, and our retriever collected references.
Additionally, since bootstrapped samples are not always satisfying, we design corresponding correction and selection strategies to filter out a high-quality subset for real training.
All these efforts jointly help to create the \dataset, a quoted and long-formed QA dataset with 45k high-quality filtered and 83k unfiltered samples.

The dataset can be formulated as a set $\mathcal{D(Q,A,R,C)}$, where $\mathcal{Q}$, $\mathcal{A}$, $\mathcal{R}$ represents the question set, the answer set, and the reference set respectively,
$\mathcal{C} \subseteq \mathcal{Q}\times\mathcal{A}\times 2^{\mathcal{R}}$ denotes the triple set of (question, answer, valid references).

Different from free text generation, in web-enhanced QA each answer $\alpha\in \mathcal{A}$ contains quotations and thus is in the form of
\begin{equation}
    \alpha = (<s_1, \mathcal{r}_1>, <s_2, \mathcal{r}_2>, \cdots, <s_n, \mathcal{r}_n>)
\end{equation}
where $<s_k, \mathcal{r}_k>$ represents the \textit{k-th} segment in answer $\alpha$, $s_k$ is a piece of quoted text, and $\mathcal{r}_k\subset \mathcal{R}$ is a set of references that $s_k$ cites.

\subsubsection{In-context Learning Inference}~\\
We adopt a subset of questions from ELI5 train set as our $\mathcal{Q}$ and leverage a vanilla Contriever~\cite{izacard2022unsupervised} (without LLM augmentation yet) in fine-grained retrieval to produce references $\mathcal{R}$.
In this work we first try on OpenAI \texttt{text-davinci-003} API to conduct 1-shot in-context learning inference to generate quoted long-formed answers (while other LLMs such as GLM-130B~\cite{glm-130b} could be good options too).
Since the in-context learning can be volatile to input forms and prompts, we take many trails to finally determine the best bootstrapping strategies as follows:

\vpara{Prompt Formulation.}
Since we input many contents to the API, including a few of demonstrations (i.e., high-quality samples ($q_d$, $\alpha_d$, $\mathcal{R}_d$)), the question, and the corresponding references, their formulation could impact the performance significantly.
We compare several types of prompts, including the order between question and its references (i.e., before or after, Cf. Figure~\ref{fig:bootstrapping} (a)), the symbols used to mark the indices of references, and the prompt words of references and questions. 
We conduct experiments with every type of prompt we have mentioned, and finally find a natural way as shown in Figure~\ref{fig:bootstrapping} (a) performs best. 

\vpara{Instruction Inducting.} 
Next, we need a proper instruction (e.g., ``Please write a answer based on the question and references.'') for guiding the LLM to generate a qualified answer.
Recent work~\cite{instruction_induction} suggests that we can take advantage of the LLM itself to design instructions for in-context learning instead of human handcrafting. 
We use several high-quality examples to induce a few possible instructions (Cf. Figure~\ref{fig:bootstrapping} (b)), and select the best-performed one based on our empirical evaluation over several queries.

\vpara{Few-shot In-Context Learning.}
We study the best shots needed for generating good quoted long-formed answers.
Because the reference parts often occupies much of sequence length, we notice that one-shot learning can surpass few-shot learning in terms of answer's quality in most time. 
Hence we finally choose to inference with one-shot demonstration sample as shown in Figure~\ref{fig:bootstrapping} (c), and finally 83k various queries and their answers have been collected.

We record the details of choosing prompts and instructions in Appendix \ref{sec:prompt-choice}.

\subsubsection{Citation Correction}~\\
We have produced a large amount of well-written quoted long-formed answers using GPT-3 in-context learning.
However, in our examination, we observe that the answers sometimes cite the wrong or invalid (i.e., nonexistent) references in their citation numbers.
As a result, to correct the citation relationships are crucial for the quality of \dataset dataset.

Despite the fact that the citation numbers can be wrong, the contents quoted in the answer are often correct. 
Thus we propose to amend the citation number according to the quotation similarity to references, by splitting an answer into few segments by generated citation numbers and match then to references.
For a question $q$, our retrieved references are defined as $\mathcal{R}$ and our answer can be defined as $\alpha$.
We define text segments $\mathcal{S}=\{s_1, s_2,\cdots, s_n\}$, and for each pair $(s, \mathcal{r})\in \mathcal{S}\times\mathcal{R}$, we compute citation match scores $f(s, r)$ for $r\in\mathcal{R}$. 
We pick a threshold $T$, and the final citation $r$ for each segment $(s, \mathcal{r})\in \alpha$ can be described as:

\begin{displaymath}
\mathcal{r}_i = \{r |f(s_i, r)\ge T \}, r\in\mathcal{R}
\end{displaymath}

For our application, we finally adopt Rouge-1 score as the $f$ and the threshold $T$ selection is introduced in the Section~\ref{sec:filtering}.

\subsubsection{Filtering}\label{sec:filtering}~\\ 
After correction, we further investigate more issues that could potentially influence the dataset quality.
And in short, we discover that most of them are related or could be solved via checking the citation quality.
We will discard a piece of generated sample if it presents any problems in the following:

\begin{itemize}[leftmargin=*,itemsep=0pt,parsep=0.2em,topsep=0.2em,partopsep=0.0em]
    \item \textbf{Hallucination~\cite{survey_hallucination}}: the answer leverages the internal knowledge of LLMs instead of references, which is not factual-grounded and sometimes severely wrong. It can be identified via the low overlapping ratio between all references and the answer.
    \item \textbf{Few citations}: when an answer cites too few of the provided references, it usually presents poor reference relevance and thus often not informative and factual-grounded enough.
    \item \textbf{Low citation accuracy}: if an answer have too many wrong citation numbers, we assume it as a low-quality one.
\end{itemize}

We calculate the F1 for the similarity and overlapping calculation. 
We test Rouge-L (whose best threshold is 0.4) and Rouge-1 (whose best one is 0.57) on a set of manually checked samples,  and find that Rouge-1 is better.
It is due to the fact that LLMs would often rewrite and paraphrase the reference contents including exchanging phrase orders. 
In that case, a high-quality answer may hold a high informative Rouge-1 score, but a low Rouge-L score, which computes the longest common subsequence co-occurrence.

After all the filtering conditions mentioned above, the number of samples drops from 83k to 45k, which becomes a high quality quoted long-formed QA dataset for web-hanced QA system training.
We train the GLM~\cite{du2022glm}, a type of bidirectional LM that is pre-trained on autoregressive blanking infilling (including a 10-billion-parameter and a 2-billion-parameter one), over the \dataset as our backbone generator.

\subsection{Human Preference-aware Scorer}
\label{sec:preference_scorer}
In preliminary testing, our bootstrapped generator under beam-search decoding strategy already performs satisfyingly in many cases.
However, recent literature~\cite{learning-to-summarize,webgpt,instructgpt} demonstrates that aligning human purposes and preference to LLMs are crucial for expert-level text generation.
WebGPT reports to recruit many experts to provide comparison and ranking over generated answers and make use of the feedback to train a reward model (RM) for picking best-of-n (i.e., 16/32/64) generated candidates and additionally optimize the generator via reinforcement learning (RL).

Nevertheless, such expert annotations could be expensive to acquire and the RL would consume much computation resource.
In this work, as a competitive substitute, we propose to build a human preference-aware scorer based on massive user feedback (e.g., thumb-ups) from online QA forums.
Under appropriate designs and elaborate data cleaning, we show in our experiments that such scorer also significantly improve the alignment-level of answers and the scoring in real human evaluation.

\vpara{Data collection and preprocessing.}
We first collect QA pairs and corresponding user thumb-ups from online QA forums.
Despite their diversity, these answers are of so various lengths and qualities that the scorer would learn little from them without proper preprocessing.

Our preprocessing includes the following requirements:
\begin{itemize}[leftmargin=*,itemsep=0pt,parsep=0.2em,topsep=0.2em,partopsep=0.0em]
    \item \textbf{High quality feedback}: we define the answer with more than 3 thumb-ups as an answer with valid feedback. 
    We pick out questions with 8 or more valid answers as qualified ones.
    \item \textbf{Length-bias mitigation}: we notice that the score prefers longer answers rather than the better ones in preliminary study, as is also indicated in literature~\cite{learning-to-summarize,instructgpt}. 
    To mitigate the bias, for each qualified question, we use the median length $x$ of all the answers as the threshold to truncate longer answers and discard those lengths are less than $x/2$.
    \item \textbf{Contrast augmentation}: after sorting the answers by their thumb-ups, the gaps between neighboring answers turn out narrow. 
    Scorers trained on such uninformative dataset present poor performance.
    To increase the contrast between answers for comparison training, we select a pair of answers of more than 5 in rank positions. In each pair, the answer with greater amount of likes is the better response.
\end{itemize}

After our prepossessing, there are 93k questions and 249k comparison pairs in total, with 230k pairs as the training set and 19k pairs as the test set.
Next, we introduce the implementation details for training our human preference-scorer.
The backbone model for training scorer is a 6-billion-parameter GLM.

\vpara{Supervised fine-tuning (SFT).}
In SFT step, we leverage the Reddit TL; DR dataset for first fine-tuning the scorer following~\cite{learning-to-summarize}. 
We train 16 epochs with cosine learning rate decay and 2.83e-5 as beginning learning rate. 
We use the SFT model for initialization of comparison training.

\vpara{Comparison training.}
We pass pairs of comparison data to the model to yield a scalar score for each of the question-answer pair and maximize the gap between their scores.
We use a linear head with the input dimension of hidden size and the output dimension of 1 to produce the score.

During the training, we find that the scorer tends to overfit quickly. 
Therefore, we freeze first 70\% transformer layers and leverage other techniques such as dropouts and large batch size for regularization. 
Notwithstanding, the scorer would overfit after 1-1.5 epochs anyway. 
After the training completes, we calibrate its predictions to standard normal distribution based on the training set reward distribution.

%% file: tables/natural_adoption.tex
\setlength{\columnsep}{10pt}
\begin{wraptable}{r}{3.75cm}
    \vspace{-3mm}
    \centering
    \caption{Evaluation on LLM's reference adoption.}
    \label{tab:natural_adoption}
    \vspace{-2mm}
    \begin{tabular}{ll}
    \toprule[1.2pt]
    Method         & Acc.  \\ \midrule
    Contriever     & 68.6\% \\
    LLM ICL adoption & 90.2\% \\ \bottomrule[1.2pt]
    \end{tabular}
    \vspace{-4mm}
\end{wraptable}

%% file: sections/6_metrics.tex
\section{Human Evaluation Criteria}
\label{sec:human_eval}

Automatic metrics to score model-generated answers can perform well in terms of short-formed ones. 
However, for open-domain long-formed QA with references, the answers and rationales can be subjective and versatile, especially for those questions that start with "HOW" and "WHY."
As a result, human evaluation is vitally needed, for which there have been many studies~\cite{survey_nlg_metrics_2006,survey_nlg_metrics_2008}. 

To evaluate \model and appropriately compare it to other similar models, we introduce a human evaluation criteria system to evaluate both references and answers.
We adopt both binary (for those objective metrics, e.g., truthfulness) and four-level score (for those subjective metrics, e.g., fluency) balancing objectivity and scale in human evaluation.
The four-level score is applied as is suggested in the literature that it avoid human annotators to keep absolutely neutral~\cite{survey_nlg_metrics_2008}.
For each criterion we mention below, an arrow follows. up arrow $(\uparrow)$ means higher score performs better, while down arrow $(\downarrow)$ denotes lower score performs better.

\subsection{Reference Evaluation}

In this section, we introduce human evaluation criteria on references.
The evaluation is done on per question-reference pair.

\vpara{Relevancy ([0, 3], $\uparrow$).}
For retrieved documents or references related to a question, the more related, the higher relevancy score should be. Specifically, different references to a question can share high relevancy scores simultaneously. 
             
\vpara{Density ([0, 3], $\uparrow$).}
To evaluate how much useful information is in a piece of reference, we need to estimate its information density. 

Both relevancy and density are criteria to evaluate informativeness, but there is difference between them. 
Relevancy can be regarded as a "recall metric" for informativeness, while density can be regarded as a "precision metric".

\vpara{Truthfulness ([0, 1], $\uparrow$).}
Retrieved references can be factually wrong even they are closely associated to the question.
It is because the web information sources are open and could contain user-submitted information without correctness check. 
As a result, the truthfulness of a piece of reference should be evaluated, and its evaluation does not consider the question.

\vpara{Toxicity ([0, 1], $\downarrow$).}
Web texts could involve violent, pornographic, offensive words or other improper elements. 
Thus, it is necessary to assess toxicity of references retrieved.

\vpara{Social Bias ([0, 1], $\downarrow$).}
Potential biases on the internet could related to genders, races, nations, and ages. 
We should also exclude them from our system.

\subsection{Answer Evaluation}
In this section, we introduce human evaluation criteria on answers, which are evaluated triple-wise (i.e., (question, answer, references)).

\vpara{Fluency ([0, 3], $\uparrow$).}
Fluency measures the quality of generated text itself only, without taking questions and references into account~\cite{survey_nlg_metrics_2006}. 
It concerns only elements such as grammar, word, and phrase choices that are affiliated to the language aspect.

\vpara{Correctness ([0, 3], $\uparrow$).}
Correctness measures the coherence of the answer and its corresponding question. 
If an answer solves the question satisfyingly, we say it holds a high correctness. 
Additionally, when we score the correctness of an answer, we should take factual consistency into account. 
For example, contradicting common sense or defying logic will decrease the correctness.

\vpara{Citation Accuracy ([0, 3], $\uparrow$).}
The metric only considers the relationships between an answer and its references.
When an answer contains citation marks, we should check if it is correct. 
Citation mistakes or missing citation will both decrease the accuracy.

\vpara{Truthfulness ([0, 1], $\uparrow$).}
Similar to truthfulness in the reference evaluation, truthfulness of an answer measures whether the text of the answer is factually sound, including the factual consistency of the answer and whether the answer contains contradictions or hallucinate information. 

\vpara{Objectivity ([0, 1], $\uparrow$).}
The metric only concerns the relationships between an answer and its references.
When references provided, models are supposed to generate answers according to these references without its using its latent knowledge from pre-training. 
If we can find all the information of an answer from provided references, we say it is objective. 

\vpara{Redundancy ([0, 1], $\downarrow$).}
Within the limited text length, duplicate content will reduce informativeness.
As the lower redundancy, the higher quality of the answer, we take it into our consideration.

The detail of the metrics and the meaning of the score can be found in the Appendix~\ref{sec:metric_details}.

%% file: sections/8_evaluation.tex
\section{Experiment}
In this section, we conduct experiments employing the metrics mentioned in Section~\ref{sec:human_eval} to evaluate and analyze the quality of the responses generated, including those from \model and other similar systems. 
We also report quantitative ablation studies on certain components in \model.

\subsection{Main Results}
We conduct the major evaluation using the 272 questions provided on WebGPT~\cite{webgpt} demo website\footnote{\url{https://openaipublic.blob.core.windows.net/webgpt-answer-viewer/index.html}}, as the WebGPT is not publicly available and selected questions are generally complicated and closer enough to real human questions. 

\vpara{Human Evaluation Setup.}
We recruited 15 master-degree level experts to conduct human evaluation. 
For each question, we aggregate all the search results and answers from different models into one table, enabling the annotators to effectively compare them and unify the annotation standards.  
We evaluate the performance of our model and other different models from various dimensions through human evaluation. We also compare and analyze the results from different perspectives as follows. The main results are shown in Table \ref{tab:main-result}.

\input{tables/main_results.tex}

\vpara{WebGLM Reference vs Other References.}
Although the search results of WebGLM are slightly inferior to WebGPT-175B, its performance is far better than that of Perplexity.ai and WebGPT-13B. It is worth mentioning that the WebGLM retrieval process only uses some traditional, word-based algorithms and two Contrievers with a cumulative parameter amount of no more than 300M. WebGLM is significantly superior to WebGPT in computing performance and time consumption. Its performance is far superior to that of the 13B model and close to that of the 175B model.

\vpara{WebGLM vs Other Systems.}
Finally, we compare our system with the results of WebGPT-13B, Perplexity.ai, and WebGPT-175B. Our system has achieved the highest performance in fluency, truthfulness, and redundancy. At the same time, we are close to WebGPT-175B in the correctness metric with a score of 2.81, which is far higher than that of Perplexity.ai and WebGPT-13B, indicating that our system can still achieve superior performance at a lower cost.

\subsection{Turing Test}
To further compare our performance, we design a Turing test~\cite{turing} to check the answers' quality.

\vpara{Setup.} We randomly sampled 200 items from the 272 questions that WebGPT has displayed on their official web page. For each question, we shuffle the answers generated by \model, WebGPT-175B, WebGPT-13B, and Perplexity.ai, and remove citation marks from them for fairness. We next mix an answer written by humans into these answers and ask evaluators to rank the answers by their quality, such as correctness, informativeness, and truthfulness. 

\begin{figure}[ht]
    \centering
    \vspace{-4mm}
    \includegraphics[width=0.85\linewidth]{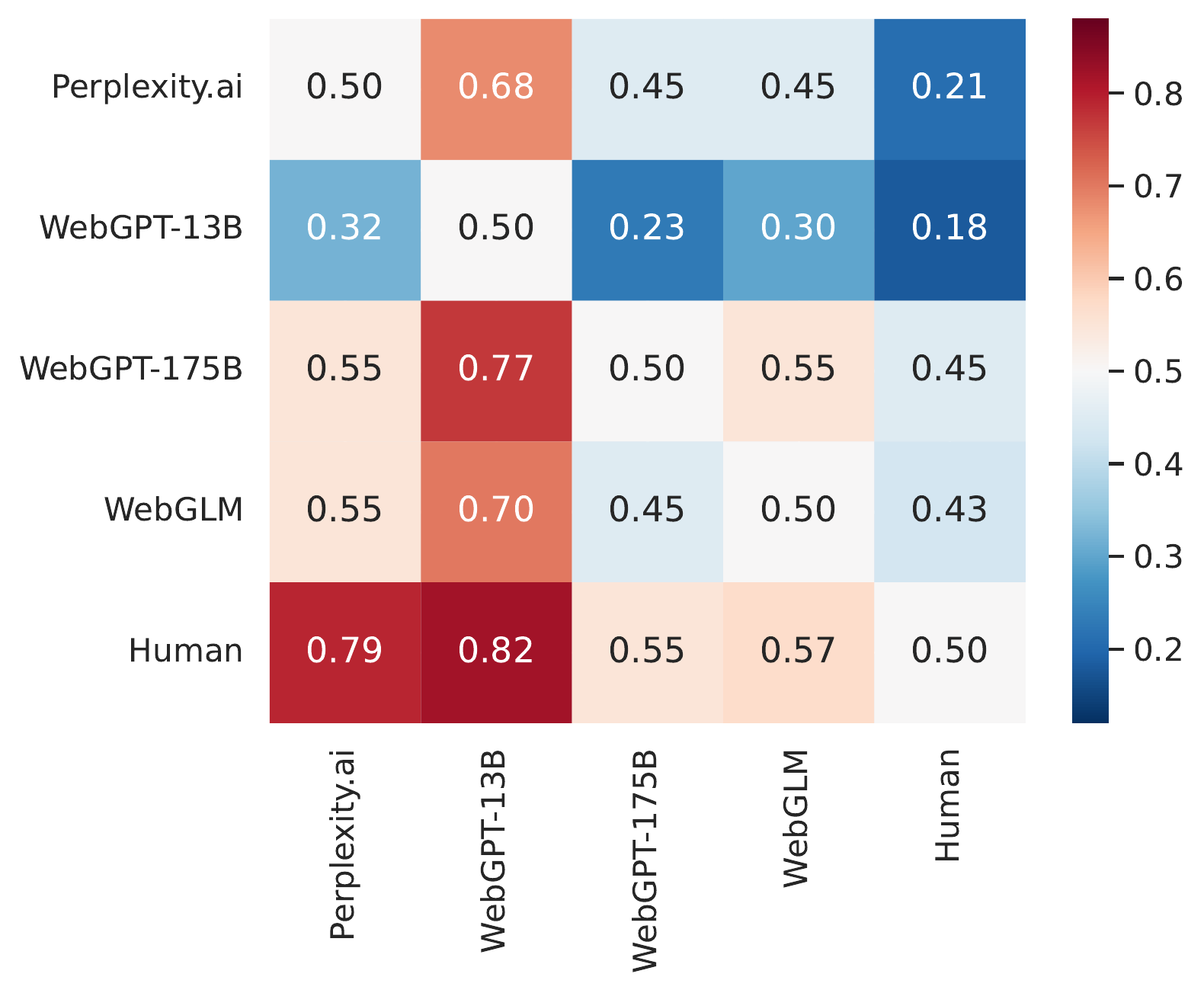}
    \vspace{-2mm}
    \caption{Win rates between systems. \textmd{Numbers denote the rate that the answers from corresponding source from the first column are better than ones from corresponding source from the first row. } }
    \vspace{-3mm}
    \label{fig:win-rate}
\end{figure}

\vpara{Result.} For each pair of answers $(A, B)$, if evaluators prefer $A$ to $B$, we call $A$ wins and $B$ loses. Firstly, we compare each pair of the answers, the win rate is shown in Figure \ref{fig:win-rate}. Besides, We calculate the win rates against humans for each system. The result is shown in Figure \ref{fig:intro_figure}. We hold a 43\% win rate, definitely beat Perplexity.ai with a 21\% win rate and WebGPT-13B with an 18\% win rate, and almost draw with WebGPT-175B with a 45\% win rate.

\subsection{Test on QA Benchmarks}

We randomly sample 400 questions on Natural Question and Web Question, and evaluate WebGLM and Perplexity.ai on them. The results in Table \ref{tab:traditional-benchmark} show that WebGLM outperform Perplexity.ai.

\input{tables/traditional_benchmark}

In addition, we conducted experiments on the full validation split of TriviaQA (same as WebGPT). Following the testing method employed by WebGPT, we first generated a long answer for each question using WebGLM. We then used Google Bigbird, fine-tuned on the TriviaQA training set\footnote{\url{https://huggingface.co/google/bigbird-base-trivia-itc}}, to answer TriviaQA questions based on the output of WebGLM. To address potential test-train overlap issues mentioned in WebGPT, we also conducted TriviaQA tests on different train-test splits. The results are summarized in Table~\ref{tab:triviaqa-benchmark}.

\input{tables/triviaqa_benchmark}

\subsection{Ablation Study}
In this section, we study the major improvements and strategies in \model, including the bootstrapped dataset filtering, scorer training, LLM-augmented retriever and some other minor topics.

\input{tables/dataset_analysis}

\subsubsection{\dataset Filtering}\label{sec:dataset_analysis}~
Since we build our training dataset based on LLM in-context bootstrapping, the dataset quality could be essential for \model's success.
We randomly sample 210 examples from these versions of our dataset to verify the filtering strategies they are based on, including 1) None, 2) Rouge-L filtered, and 3) Rouge-1 filtered.

We randomly shuffle all the samples and distribute them to evaluators, and then collect and calculate the average score of each metric. The sample results are shown in Table~\ref{tab:dataset-analysis}
We analyze this result from two perspectives. 
One is the absolute performance of our final version of the dataset.
The other is comparing the performance of our different versions of datasets.

We find that our dataset holds a high factual consistency and correctness, and the majority of our data are judged as perfectly correct.
We have also noticed that the information relevancy and density are considerably improved when we apply a filter method and when we change Rouge-L to Rouge-1.
As for the answer, we find that correctness has great improvement when we apply any one of the two filters, and factual consistency has a great improvement when we change the Rouge-L filter to Rouge-1. Besides, objectivity is also one of the most important criteria that we care about, and we find that it's more likely to discard subjective answers with a Rouge-1 filter than with a Rouge-L filter.
As a result, our experiments show that citation accuracy is closely related to the reference quality and answer quality, so our filter method is effective.

Besides, We train the GLM-2B models on each dataset and evaluate them with our designed metrics to see the impact of these datasets on our model's performance. We show the results in Table~\ref{tab:ref_based_2b_humean_eval}. We find that the answers of the three models showed little difference in the correctness metric. However, the performance of the model trained by rouge-1 was better in fluency, citation accuracy, and objectivity metrics. This result further proves the advantages of the dataset of rouge-1. Therefore, we decide to train our 10B model on the dataset of rouge-1.

\input{tables/ref_based_2b_human_eval.tex}

\subsubsection{LLM-augmented Retriever}
In terms of the usefulness of references, we have compared our method with traditional methods such as BM25, TF-IDF, and the original version of Contriver. 

We collect 22000 examples from WebGLM-QA, and for each question, we calculate Rouge-1 precision score $p$ of corresponding answer $a$ and each of the reference $r$, and then label the reference-answer pair $(r, a)$ as $p$. Finally, we gain a training dataset containing 20000 examples and a test dataset containing 2000 examples.

\input{tables/contriever_v2_performance.tex}

For all answers to the same question, we compare the order predicted by retrieve methods with the answer relevancy order. 
The results are shown in Table~\ref{tab:contriever-2-performance}.
We notice that before the LLM task augmentation, the Contriever performs even poorer than traditional lexical-based approaches.
After augmenting knowledge from GPT-3's reference adoption labeling, we find that ours, which holds a 69.36 pair-wise choosing accuracy and 62.26 spearman index, performs best.
The evidence strongly advocates that the LLM augmentation is vital when we use pre-trained smaller dense retrievers in practice.

\subsubsection{Human Preference-aware Scorer}
In this section we compare several different scorer training strategies and datasets.
We discover that proper task formulation and larger and more diverse dataset yield better results.

\vpara{Baseline and data preprocessing.}~
We first train RoBERTa-large in the classification task and the regression task formulation, and the 6-billion-parameter GLM on the ELI5's training set (with thumb-ups) as our baselines.
In the classification task, we collect all items whose count of answers is not less than 10 from ELI5. For each collected question, we label top-5-voted answers as positive, and randomly pick 5 answers from other questions as negative examples. 
In the regression task, we collect all items whose count of answers is not less than 5 from ELI5. For each collected question, we complete the following steps: 

\begin{enumerate}[leftmargin=*,itemsep=0pt,parsep=0.2em,topsep=0.2em,partopsep=0.0em]
    \item for each answer to this question, supposing its corresponding up-vote is $u$, we firstly label this answer as $\log_2{(u+1)}$. 
    \item Then, we scale labels of all answers to this question to $[0,1]$. 
    \item Let $x$ be the summation of the answers' label, we randomly pick $\lfloor x\rfloor$ answers from other questions as negative examples with label $-1$. 
\end{enumerate}

\input{tables/ablation_study}

In order to obtain a large train set (which has been suggested very important in~\cite{learning-to-summarize}), we adopt a relatively loose screening method, which selects the questions with more than 5 answers and answers with no less than 100 words in length. Our large train set includes 28.2k questions and 191.6k pairs. We use the ELI5 test set with thumb-ups for our final evaluations.

\vpara{Metrics.}
We select three metrics to measure the ability of the reward model to distinguish responses of different quality, namely accuracy, Spearman coefficient, and NDCG (Normalized Discounted Cumulative Gain). Accuracy refers to the accuracy of selecting better answers in pairs. Spearman and NDCG measure the sorting ability of the model.

\input{tables/reward_compare}

The ranking evaluation of different models is shown in Table~\ref{tab:reward_compare}. We find that WebGLM human preference-aware scorer performs best on accuracy and Spearman coefficient. Under the same amount of training tokens, the performance of the reward model is slightly worse than that of RoBERTa classification and RoBERTa regression, but after increasing the amount of training, the performance of the reward model will increase significantly.

\begin{figure}[t]
    \centering
    \vspace{-1mm}
    \includegraphics[width=0.95\linewidth]{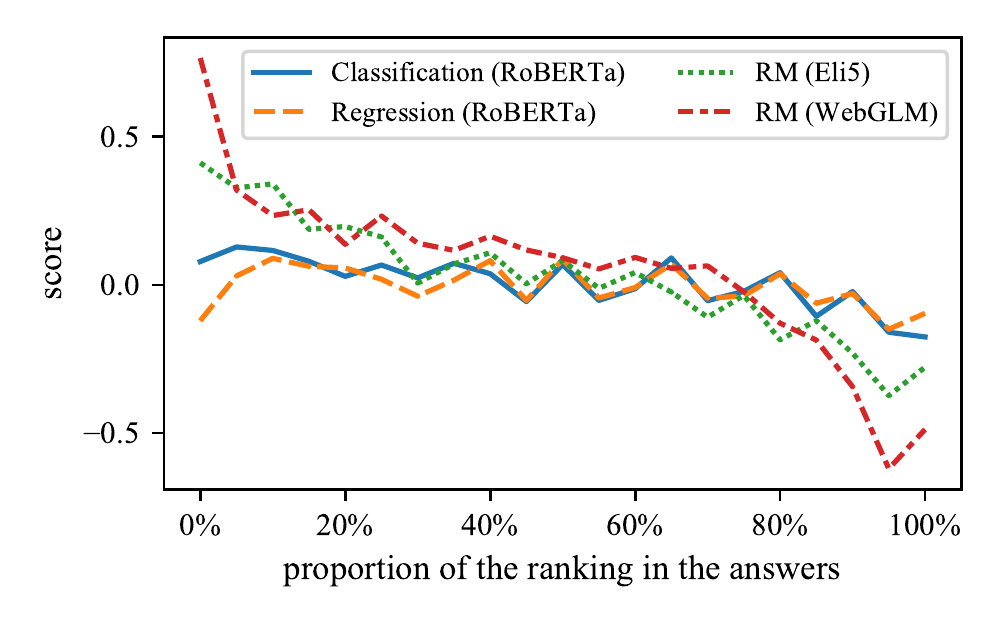}
    \vspace{-3mm}
    \caption{Average score of answers in ELI5 test set. \textmd{It is sorted by likes in the ELI5 test set. The best answer is around $0\%$ and the worst answer is around $100\%$.} } 
    \label{fig:reward}
    \vspace{-6mm}
\end{figure}

Figure \ref{fig:reward} shows the average reward of the answers at different positions in the sequence sorted by likes in the ELI5 test set. The best answer is around $0\%$ and the worst answer is around $100\%$. We find that the curve of the WebGLM Human Preference-aware Scorer is more discriminative than other models, and the rewards of the best answer are higher than that of others.

\subsubsection{Ablation Study on Each Component}
We added some experiments to conduct ablation studies on each component. We compared the three sub-modules of the system: Retriever, Generator, and Scorer. The results are shown in Table~\ref{tab:ablation_study}.

In the Retriever module, we compared the performance on the settings of WebGPT-175B, WebGLM, and non-retrieval. From the Table~\ref{tab:ablation_study}, the performance on WebGLM retrieval is similar to that of WebGPT-175B and significantly better than non-retrieval.

Regarding the Generator module, we compared the response quality of WebGLM and GPT-3 on WebGLM retrieval setting. We found that WebGLM performed slightly better than GPT-3 in fluency, correctness, accuracy, citation accuracy, objectivity, and truthfulness.

In terms of Scorer, we compared the response quality of WebGLM removing and retaining Reward Models. The results show that by WebGLM-10B top-p sampling and reward model scoring method, We found through the human evaluation results that the answers scored high by the reward model excel the original results in fluency, correctness, citation accuracy, truthfulness, and redundancy. It shows the importance of the reward model scoring mechanism to model performance.

%% file: tables/main_results.tex
\begin{table*}[ht]
\caption{Main results based on human evaluation metrics. \textmd{Human evaluation results of generations on questions provided on the WebGPT demo website. For reference evaluation, Rel., Den., Tru., Tox↓., and Soc. Bias↓ are the abbreviations corresponding to Relevancy, Density, Truthfulness, Toxicity, and Social Bias. For answer evaluation, Flu., Cor., Cit. Acc., Obj., Tru., Red.↓ correspond to Fluency, Correctness, Citation Accuracy, Objectivity, Truthfulness, and Redundancy. }}
\vspace{-2mm}
\label{tab:main-result}
\renewcommand\tabcolsep{7pt}
\begin{tabular}{@{}lccccccccccc@{}}
\toprule
\multirow{2}{*}{Model} & \multicolumn{5}{c}{Reference Evaluation}                                           & \multicolumn{6}{c}{Answer Evaluation}                                                               \\ \cmidrule(l){2-6} \cmidrule(l){7-12}
                       & Rel.           & Den.           & Tru.           & Tox.↓          & Soc. Bias↓      & Flu.           & Cor.           & Cit. Acc.      & Obj.           & Tru.           & Red.↓          \\ \midrule
WebGPT (175B)          & 2.512          & 2.660          & 0.996          & 0.015          & 0.006          & 2.457          & 2.889          & 2.837          & 0.990          & 0.975          & 0.087          \\ \midrule
Perplexity.ai          & 1.652          & 1.636          & 0.955          & \underline{0.005}          & \textbf{0.001} & \underline{2.718}    & \underline{2.321}    & 2.512          & 0.726          & \underline{0.975}    & \underline{0.032}    \\
WebGPT (13B)           & \underline{1.782}          & \underline{1.766}          & \textbf{0.998} & 0.008          & 0.016          & 2.692          & 2.102          & \textbf{2.769} & \textbf{0.974} & 0.872          & 0.051          \\
WebGLM (10B)           & \textbf{1.980} & \textbf{2.226} & \underline{0.983}    & \textbf{0.002} & \underline{0.002}    & \textbf{2.829} & \textbf{2.810} & \underline{2.757}    & \underline{0.943}    & \textbf{0.998} & \textbf{0.021} \\ \bottomrule
\end{tabular}
\vspace{-4mm}
\end{table*}

%% file: tables/traditional_benchmark.tex
\begin{table}[ht]
\caption{Open QA Performance on NaturalQuestions and WebQuestions. \textmd{Perplexity.ai is evaluated on sampled subsets because the website prohibits crawling.}}
\label{tab:traditional-benchmark}
\centering
\vspace{-2mm}
\renewcommand\tabcolsep{5pt}
\begin{tabular}{@{}ccccc@{}}
\toprule
                           & Natural Questions& Web Questions \\ \midrule
WebGLM                     & \textbf{60.8}   & \textbf{63.5}               \\
Perplexity.ai (sample)        & 57.3   & 57.5    \\
GPT3-175B                  & 29.9       & 41.5     \\ \bottomrule
\end{tabular}
\vspace{-3mm}
\end{table}

%% file: tables/triviaqa_benchmark.tex
\begin{table*}[ht]
\caption{WebGLM, WebGPT and other comparison methods on TriviaQA. \textmd{The setting follows WebGPT~\cite{webgpt} Appendix G.}}
\label{tab:triviaqa-benchmark}
\centering
\vspace{-2mm}
\renewcommand\tabcolsep{8.5pt}
\begin{tabular}{@{}lcccccc@{}}
\toprule
Method                      & Total            & \makecell[c]{Question\\overlap} & \makecell[c]{No question\\overlap} & \makecell[c]{Answer\\overlap}   & \makecell[c]{Answer\\overlap only} & No overlap       \\ \midrule
Bigbird + WebGLM (Ours)     & \textbf{70.80\%} & 86.40\%          & \textbf{67.10\%}    & \textbf{78.70\%} & \textbf{73.60\%}    & 49.30\%          \\
GPT-3 175B                  & 58.70\%          & 75.90\%          & 52.90\%             & 67.30\%          & 61.60\%             & 39.00\%          \\
GPT-3 175B + WebGPT 175B BC & 69.50\%          & 86.30\%          & 65.30\%             & 78.40\%          & 73.20\%             & \textbf{52.40\%} \\
UnitedQA-E                  & 68.90\%          & \textbf{89.30\%} & 62.70\%             & 78.60\%          & 70.60\%             & 44.30\%          \\
UnitedQA (hybrid model)     & 70.50\%          & -                & -                   & -                & -                   & -                \\ \bottomrule
\end{tabular}
\vspace{-2mm}

\end{table*}

%% file: tables/dataset_analysis.tex
\begin{table*}[ht]
\caption{Ablation study on different dataset filtering strategies in creating the bootstrapped generator.}
\label{tab:dataset-analysis}
\vspace{-2mm}
\renewcommand\tabcolsep{7pt}
\renewcommand\arraystretch{1.05}
\begin{tabular}{@{}lccccccccccc@{}}
\toprule
\multirow{2}{*}{\makecell[c]{Filtering\\Method}}  & \multicolumn{5}{c}{Reference Evaluation}                                                      & \multicolumn{6}{c}{Answer Evaluation}                                                                          \\ \cmidrule(l){2-6} \cmidrule(l){7-12}
  & Rel.           & Den.           & Tru.           & Tox.↓           & Soc. Bias↓           & Flu.           & Cor.           & Cit. Acc.       & Tru.           & Obj.           & Red.↓          \\ \midrule
None    & 1.711          & 1.619          & 0.991          & 0.011          & 0.011          & \textbf{2.872} & 2.636          & 2.370          & 2.810          & 0.805          & 0.134          \\
Rouge-L & \textbf{1.833} & 1.728          & \textbf{0.994} & 0.022 & \textbf{0.010}          & 2.731          & 2.680          & 2.573          & 2.896          & 0.841          & 0.181          \\
Rouge-1 & 1.832          & \textbf{1.751} & 0.993          & \textbf{0.010}          & 0.012 & 2.826          & \textbf{2.694} & \textbf{2.688} & \textbf{2.919} & \textbf{0.890} & \textbf{0.120}
\\ \bottomrule
\end{tabular}
\vspace{-2mm}
\end{table*}

%% file: tables/ref_based_2b_human_eval.tex
\begin{table}[t]
\caption{Ablation study on different dataset filtering strategies, based on GLM-2B's post-training evaluation}
\label{tab:ref_based_2b_humean_eval}
\renewcommand\tabcolsep{5pt}
\begin{tabular}{@{}ccccccc@{}}
\toprule
                   & Flu.                 & Cor.                 & Cit. Acc.            & Obj.                 & Tru.                 & Red.↓                 \\ \midrule
None & 2.610                & 2.738                & 2.655                & 0.961                & 0.961                & 0.063                \\
Rouge-L   & 2.604                & \textbf{2.742} & 2.727                & 0.952                & \textbf{0.975} & \textbf{0.034} \\
Rouge-1   & \textbf{2.852} & 2.738                & {\textbf{2.743}} & \textbf{0.976} & 0.970                & 0.044                \\ \bottomrule
\end{tabular}
\end{table}

%% file: tables/contriever_v2_performance.tex
\begin{table}[t]
\caption{Performance of LLM-augmented Retriever (Ours). \textmd{``N-NDCG'' refers to Normalized NDCG.}}
\label{tab:contriever-2-performance}
\renewcommand\tabcolsep{10pt}
\begin{tabular}{@{}ccccc@{}}
\toprule
   Metric(\%)           & TF-IDF & BM25   & Contriever & Ours   \\ \midrule
Accuracy       & 46.85  & 40.33  & 18.54      & \textbf{69.36} \\
Spearman        & 9.92   & -20.94 & -1.58      & \textbf{62.26} \\
NDCG            & 82.54  & 76.28  & 81.16      & \textbf{91.99} \\
N-NDCG & 46.05  & 26.77  & 41.75      & \textbf{75.29} \\ \bottomrule
\end{tabular}
\end{table}

%% file: tables/ablation_study.tex
\begin{table*}[t]

\caption{Ablation study on different sub-modules (Scorer, Retriever, and Generator) in WebGLM.}
\renewcommand\tabcolsep{12pt}
\label{tab:ablation_study}
\renewcommand\tabcolsep{14pt}
\renewcommand\arraystretch{.95}
\vspace{-2mm}
\begin{tabular}{@{}ccccccc@{}}
\toprule
Method                               & Flu.  & Cor.  & Cit. Acc. & Obj.  & Tru.  & Red.↓ \\ \midrule
\multicolumn{7}{c}{Scorer Ablation}                                                      \\  \midrule
No Scorer                                & 2.797 & 2.757 & 2.723     & 0.961 & 0.970 & 0.039 \\
Human Preference-aware Scorer (Ours) & 2.829 & 2.810 & 2.757     & 0.943 & 0.998 & 0.021 \\  \midrule
\multicolumn{7}{c}{Retriever Ablation (w.o. RM)}                                         \\ \midrule
No Retriever                       & 2.364 & 1.982 & -         & -     & 0.645 & 0.091 \\
WebGPT Retriever                    & 2.750 & 2.884 & 2.808     & 0.981 & 0.980 & 0.038 \\
Contriever                    & 2.761 & 2.732 & 2.721     & 0.963 & 0.930 & 0.043 \\
LLM-augmented Retriever (Ours)       & 2.797 & 2.757 & 2.723     & 0.961 & 0.970 & 0.039 \\ \midrule
\multicolumn{7}{c}{Generator Ablation (w.o. RM)}                                         \\ \midrule
GPT-3 (text-davinci-003, zero-shot)  & 2.751 & 2.752 & 2.607     & 0.927 & 0.966 & 0.034 \\
Bootstrapped Generator (Ours)        & 2.797 & 2.757 & 2.723     & 0.961 & 0.970 & 0.039 \\ \midrule
WebGLM (Ours)         & 2.829 & 2.810 & 2.757     & 0.943 & 0.998 & 0.021 \\ \bottomrule
\end{tabular}
\end{table*}

%% file: tables/reward_compare.tex
\begin{table}[t]
\vspace{-2mm}
\caption{Different scorers' performance on ELI5 test set.} 
\label{tab:reward_compare}
\renewcommand\tabcolsep{6pt}
\vspace{-2mm}
\begin{tabular}{@{}cccc@{}}
\toprule
                                          & Accuracy & Spearman & N-NDCG \\ \midrule
Classification (RoBERTa)         & 0.552        & 0.129          & 0.319         \\
Regression (RoBERTa)              & 0.569        & 0.164          & 0.352         \\
RM (ELI5) & 0.568        & 0.197          & \textbf{0.406}         \\
RM (WebGLM)      & \textbf{0.596}        & \textbf{0.241}          & 0.367         \\ \bottomrule
\end{tabular}
\vspace{-3mm}
\end{table}

%% file: sections/conclusion.tex
\section{Conclusion}
We build the LLM-based question-answering system---WebGLM---with a web retrieval method. 
We propose a fast and cost-effective method to retrieve valuable information from the Internet. 
We leverage GPT-3's in-context learning ability to build a LLM-bootstrapped quoted and long-form QA dataset, which is used to train our model. 
Further, we train a human preference-aware scorer and use it to give marks to responses generated by our model.
For each question, the scorer can select the highest-scored response from candidates, thus obtaining a final answer humans prefer the most. 
We conduct extensive experiments, including both the human evaluation and the Turing test, to demonstrate the competitive performance of WebGLM with some of the pioneering web-enhanced question answering systems like Perplexity.ai and WebGPT.

%% file: sections/appendix.tex
\input{sections/9_ablation.tex}

\section{Detailed Efficiency Analysis}
\label{sec:efficiency}

At the retrieval stage, we only search for one time, then take the first few results links to fetch the web pages in parallel. We then extract all paragraphs and rank these paragraphs by Contriever, and finally take the top 5 paragraphs as references. Let $t_s$, $t_f$, $t_e$, and $t_r$ denote the time we consume in four steps, so the total time we consume is $t_s + t_f + t_e + t_r$.

WebGPT simulates the operations in a virtual browser environment while obtaining references. For the 272 questions they showed, we count the types of actions and the average number of generated tokens as shown in Table \ref{tab:webgpt-175b-time} and \ref{tab:webgpt-13b-time}. Then we calculate the average time it takes to browse. Assuming that the total time $M$ ($M$ is either WebGPT-175B or WebGPT-13B) takes to generate commands in the browsing process of each question is expected to be $t_c(M)$, the time $M$ consumes $T(M)$ satisfies the following equations.

\begin{equation}
    \label{equ:webgpt-175b-time}
    T(\text{WebGPT-175B}) = t_c(\text{WebGPT-175B}) + t_s * 3.82 + t_f * 6.96
\end{equation}

\begin{equation}
    \label{equ:webgpt-13b-time}
    T(\text{WebGPT-13B}) = t_c(\text{WebGPT-13B}) + t_s * 4.05 + t_f * 7.56 
\end{equation}

We test the efficiency of GPT-3. With a 500-token prompt, the 175B model generates about 20 tokens per second, and the 13B model generates 100 tokens per second, meaning that:

\begin{equation}
    t_c(\text{WebGPT-175B}) = \dfrac{580.08\ \text{tokens/query}}{20\ \text{tokens/second}} = 29\ \text{seconds}
\end{equation}

\begin{equation}
    t_c(\text{WebGPT-13B}) = \dfrac{580.08\ \text{tokens/query}}{100\ \text{tokens/second}} = 5.8\ \text{seconds}
\end{equation}

In practice, $t_s$, $t_f$, $t_e$, and $t_r$ are about $1.81$, $2.38$, $0.29$, and $0.89$ respectively. So we consume $5.36$ seconds for one query on average. Nevertheless, assuming in the same network circumstance, the time consumption of WebGPT models can be calculated by Equation \ref{equ:webgpt-175b-time} and \ref{equ:webgpt-13b-time}.

\begin{equation}
    T(\text{WebGPT-175B}) = 52.48 \ \text{seconds}
\end{equation}

\begin{equation}
    T(\text{WebGPT-13B}) = 31.12 \ \text{seconds}
\end{equation}

Therefore, WebGPT-175B costs 52.48 seconds, and WebGPT-13B costs 31.12 seconds. Our efficiency can be about 10 times that of WebGPT-175B and 6 times that of WebGPT-13B.

\input{appendix/webgpt-time-cost}

\section{Choice of Prompts and Instructions}
\label{sec:prompt-choice}

Firstly, we attempt the zero-shot approach for bootstrapping data. To produce data with appropriate citation marks, we require specific instructions. We experiment with several methods, however, they are all limited in their effectiveness.

\texttt{Use a mark for each helpful reference you cited, such as [1].} Limitation: bootstrapped data contain mixed usage of \texttt{[1][2]} and \texttt{[1, 2]}.

\texttt{Use a mark for each helpful reference you cited, such as [1]. If there are multiple citations at one position, please use a format like [1][2][3].} Limitation: bootstrapped data contain citations of useless references.

\texttt{Use a mark for each helpful reference you cited, such as [1]. If there are multiple citations at one position, please use a format like [1][2][3]. If a reference is useless, do not cite it.} Limitation: useless references are still cited. This method do not work.

We then select few-shot context to bootstrap data. If we provide too many references or in-context examples, it is easy to exceed the token count limit. Therefore, we choose to use an 1-shot example and 5 references. We also include some useless references in the example, which are not cited in the answer.

After that, We conduct experiments on prompts and demonstrate that placing the question after the references is the most effective approach.

Regarding instruction induction for in-context learning, we experiment with the previously mentioned examples as well as some new ones, such as:

\texttt{Answer the question based on the following references with citations. Use a mark for each helpful reference you cited, such as [1]. If there are multiple citations at one position, please use a format like [1][2][3]. If a reference is useless, do not cite it.}
 
\texttt{I will provide you with some references. Based on the references, please answer my question. Pay attention that you should be objective, and you should not use your knowledge. Use a mark for each helpful reference you cited, such as [1]. If there are multiple citations at one position, please use a format like [1][2][3]. If a reference is useless, do not cite it.}

However, these instructions are too verbose, and in the presence of examples, the model's performance is not significantly impacted by the instructions. Therefore, we adopt a more natural approach to generate instructions\cite{instruction_induction} to produce a natural instruction that is interpretable by the model.

Finally, we use a very concise instruction: \texttt{Read the references provided and answer the corresponding question.}

In addition, we compared models trained with different prompt strategies, and the results are shown in the Table \ref{tab:prompt-strategy}. From the "Correctness" column, we can see the significant difference that the order of references and question in the prompt makes.

\begin{table}[ht]
\caption{The performance with training data bootstrapped by difference prompt strategies.}
\label{tab:prompt-strategy}
\centering

\resizebox{\columnwidth}{!}{

\begin{tabular}{@{}c|cccccc@{}}
\toprule
Prompt                    & Flu.  & Cor.  & Cit. Acc. & Obj.  & Tru.  & Red.  \\ \midrule
WebGLM Prompt             & 2.797 & 2.757 & 2.723     & 0.961 & 0.970 & 0.039 \\
Question before Reference & 2.633 & 2.518 & 2.700     & 0.933 & 0.970 & 0.058 \\
3-Reference               & 2.658 & 2.412 & 2.819     & 0.933 & 0.930 & 0.065 \\ \bottomrule
\end{tabular}

}

\end{table}

\section{Dataset Examples}
An example of \textbf{WebGLM-QA} is shown in Table~\ref{tab:dataset-example1}.
\input{appendix/dataset-example1.tex}

\section{Retrieval Example}
An example of retrieved references from each system is shown in Table~\ref{tab:reference-example1} and Table~\ref{tab:reference-example2}.
\input{appendix/reference-example1.tex}

\section{Answer Examples}
Some examples consisting of only answers are shown in Table \ref{tab:answer-example1} and Table \ref{tab:answer-example2}. We remove citation marks for all the answers to evaluate answers only. In this example, WebGLM's answer is consistent with the question, while the answers of two WebGPT models are beside the point in the conclusion.

\input{appendix/answer-example1.tex}

\section{Reward Model Example}

WebGLM performs better after implementing the reward model. An example is shown in Table~\ref{tab:rm-example}.

\input{appendix/rm-example.tex}

\section{Criteria Details}

\label{sec:metric_details}

The criteria details of human evaluation are shown in Table~\ref{tab:reference-criteria-details} and Table~\ref{tab:answer-criteria-details}.

\input{appendix/metric-details.tex}

\begin{figure}[ht]
    \caption{\model web demo page} 
    \label{fig:demo1_narrow}
    \includegraphics[width=\linewidth]{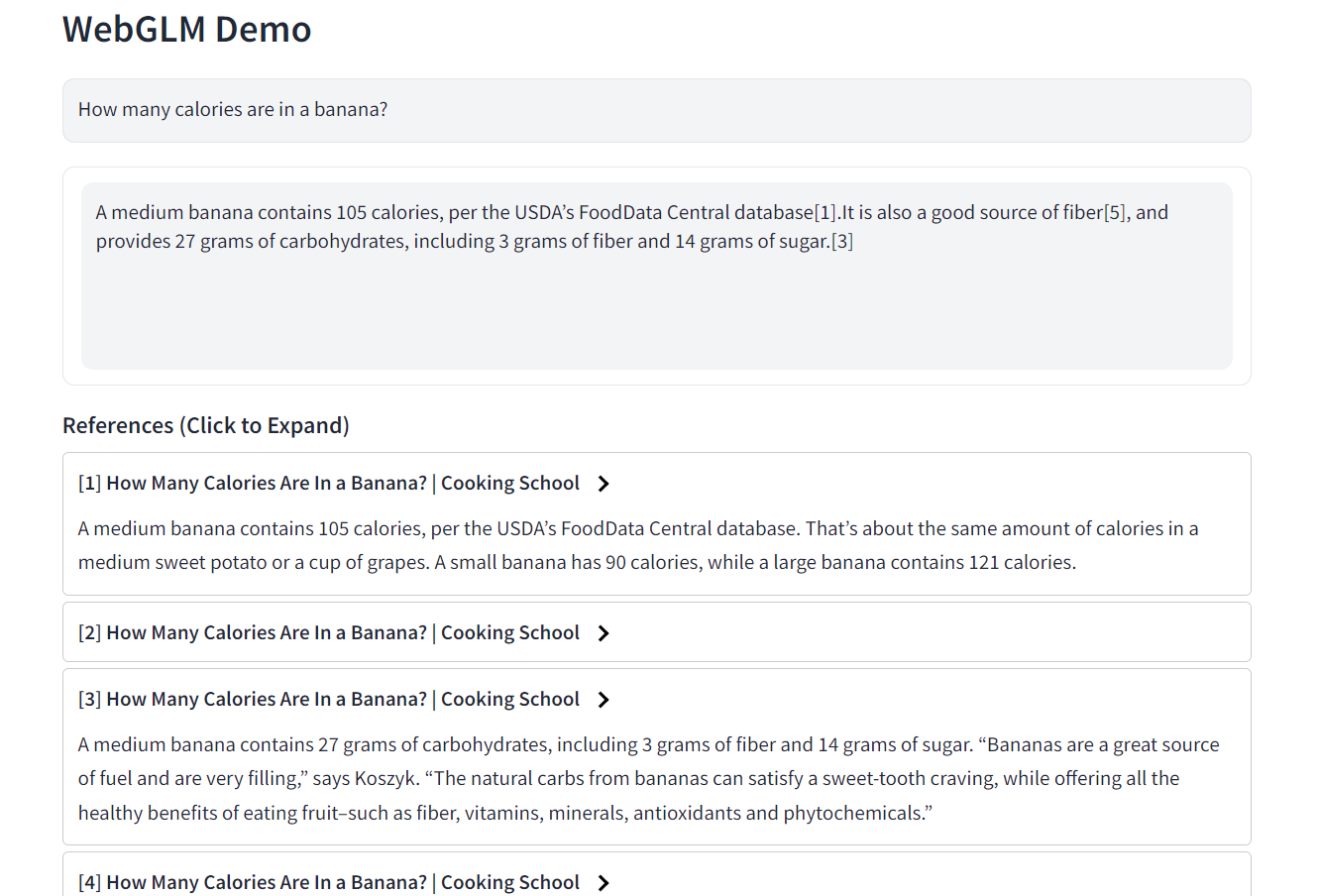}
\end{figure}

\section{Demo}

\begin{figure*}[ht]
    \caption{Real Example: How to balance career and hobbies?} 
    \label{fig:screenshot-1}
    \includegraphics[width=\linewidth]{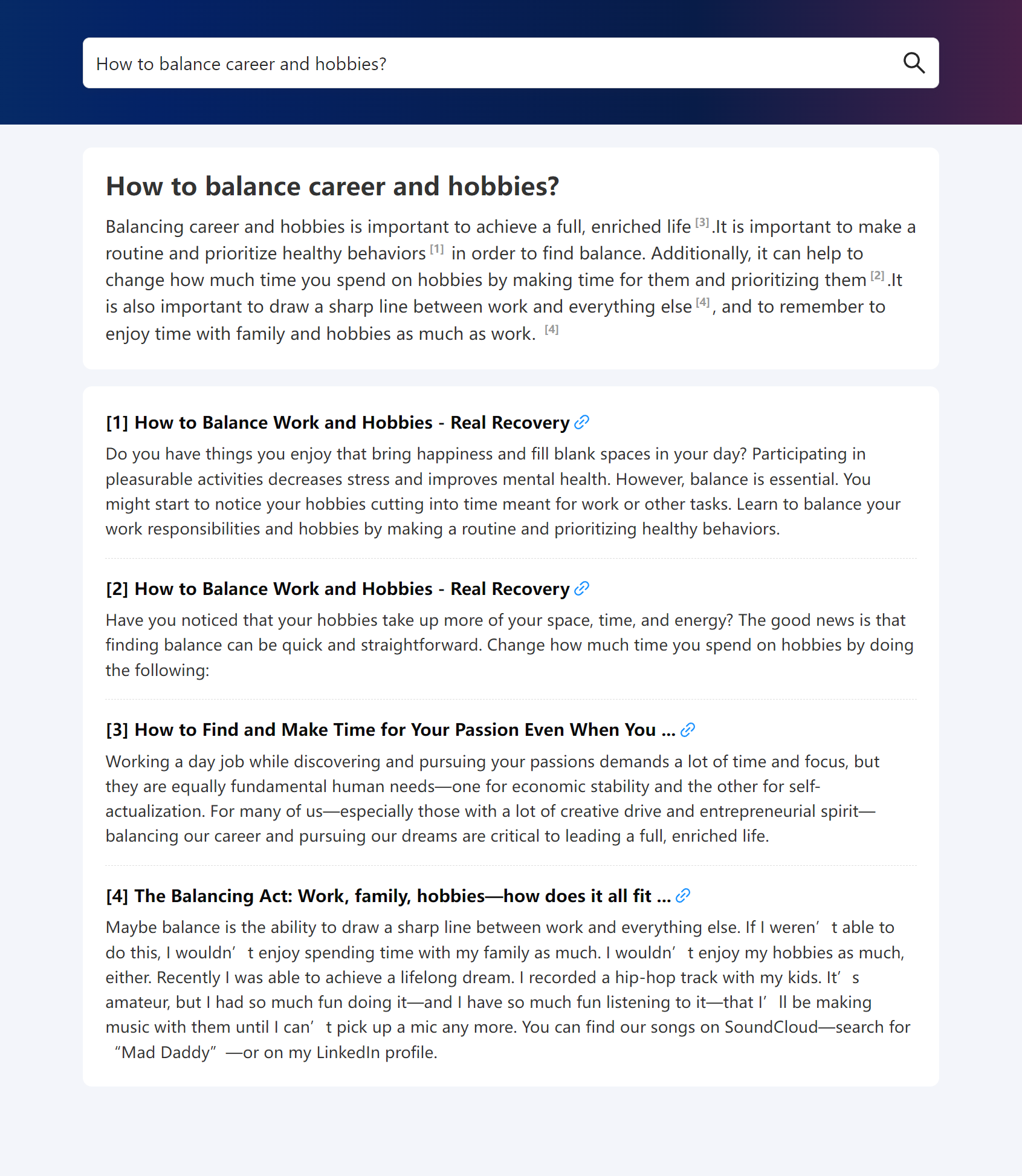}
\end{figure*}

We provide a web demo as shown in Figure \ref{fig:demo1_narrow}. Meanwhile, some real examples are shown in Figure \ref{fig:screenshot-1} and the later figures. 

\begin{figure*}[ht]
    \caption{Real Example: FL Studio and Cubase, which is better?} 
    \label{fig:screenshot-2}
    \includegraphics[width=\linewidth]{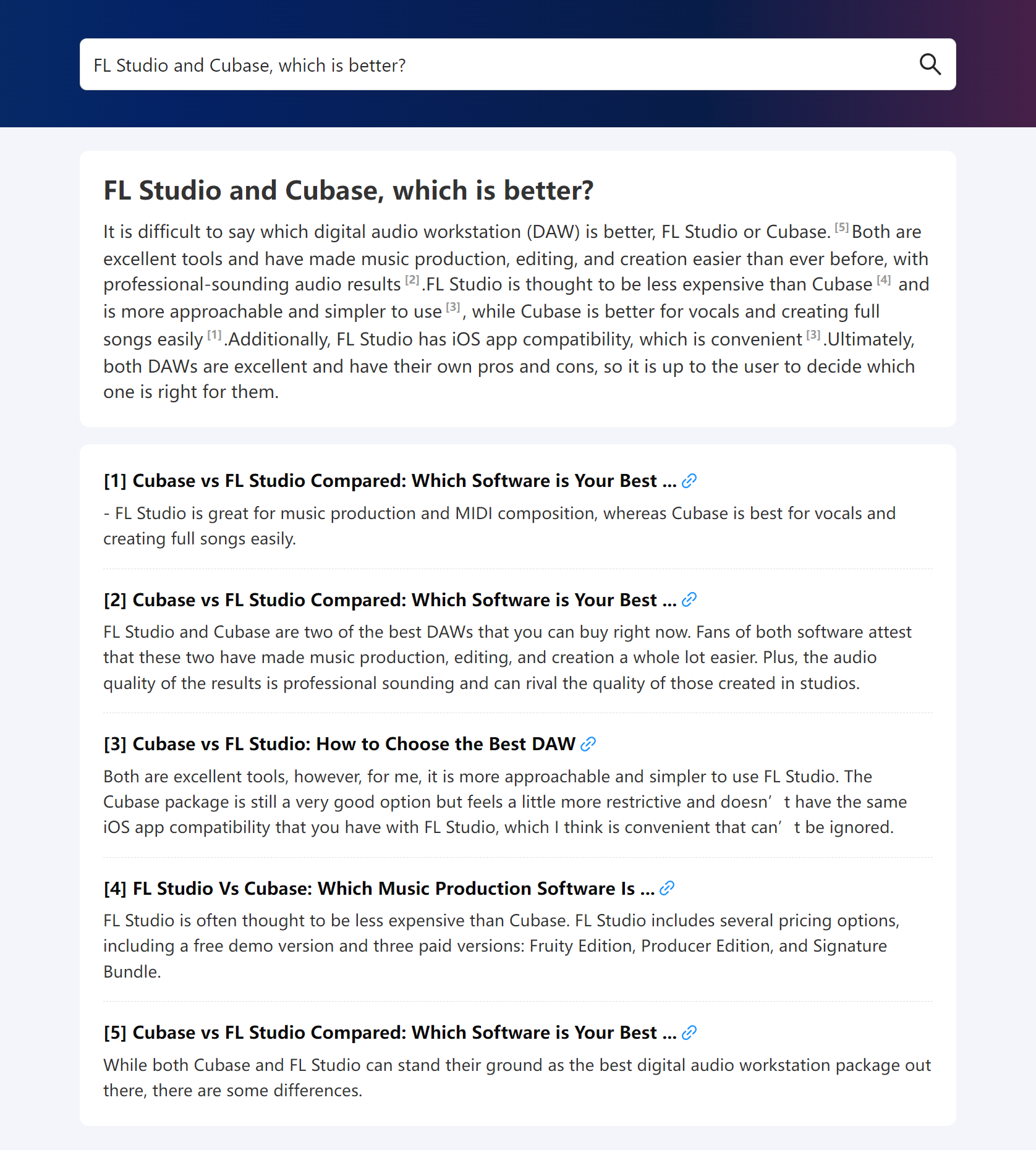}
\end{figure*}

\begin{figure*}[ht]
    \caption{Real Example: Is attention better than CNN?} 
    \label{fig:screenshot-3}
    \includegraphics[width=\linewidth]{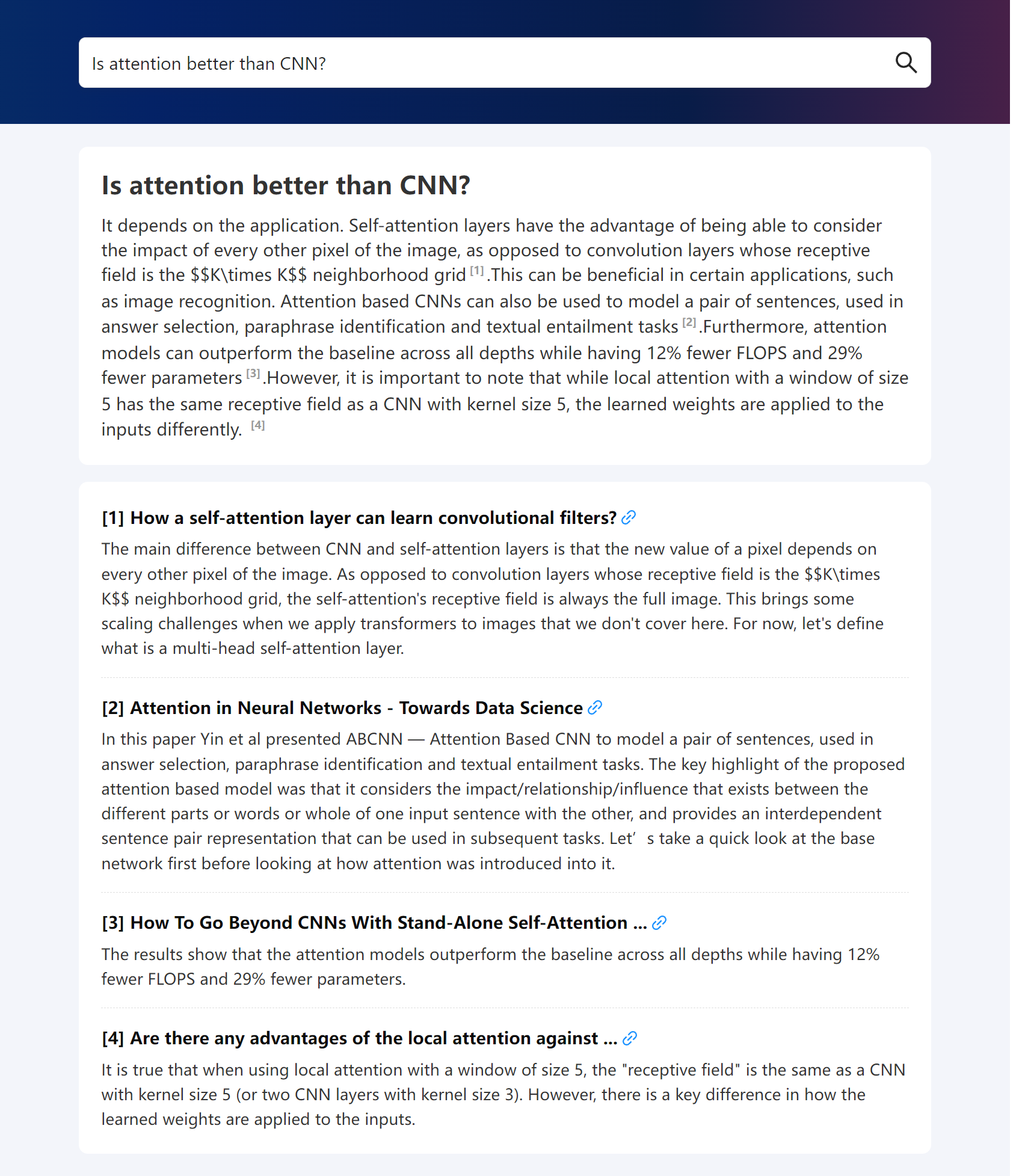}
\end{figure*}

\begin{figure*}[ht]
    \caption{Real Example: How to survive in the first-tier cities without a high-salary work?} 
    \label{fig:screenshot-4}
    \includegraphics[width=\linewidth]{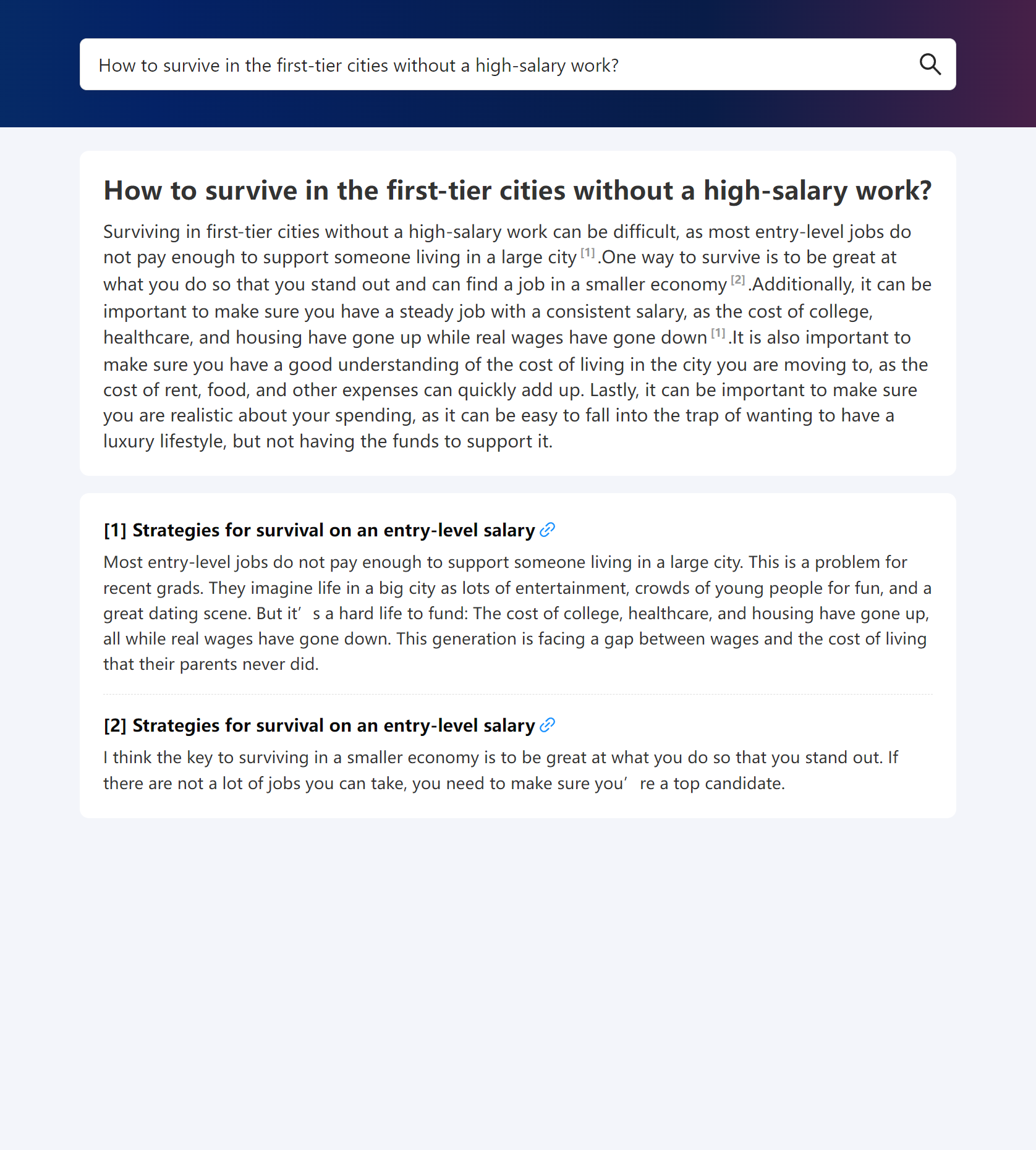}
\end{figure*}

\newpage
\begin{figure*}[ht]
    \caption{Real Example: What do you think of the 3.5 version of Genshin Impact?} 
    \label{fig:screenshot-5}
    \includegraphics[width=\linewidth]{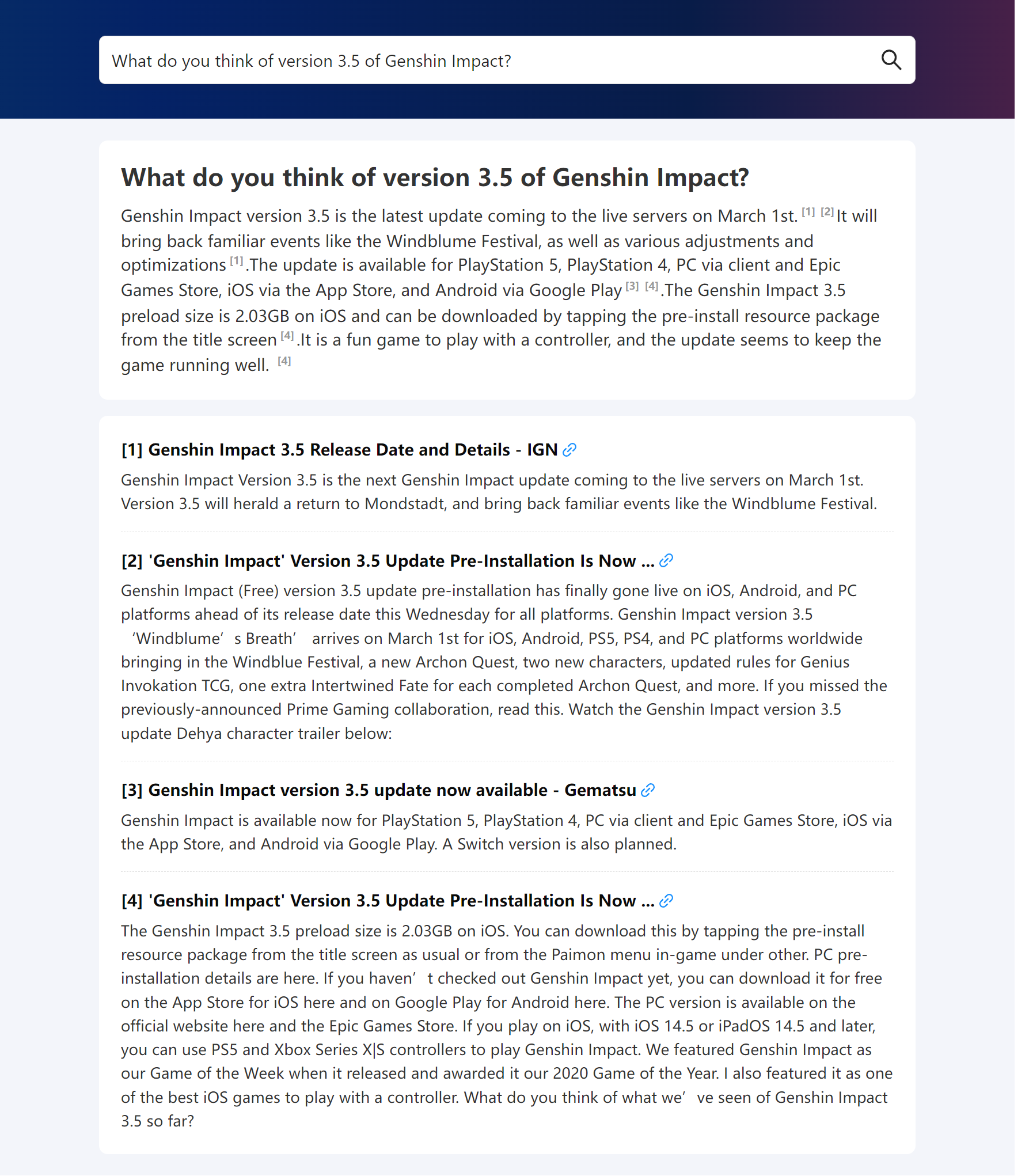}
\end{figure*}

\begin{figure*}[ht]
    \caption{Real Example: transformers are originated from NLP, but why they can be applied in CV?} 
    \label{fig:screenshot-6}
    \includegraphics[width=\linewidth]{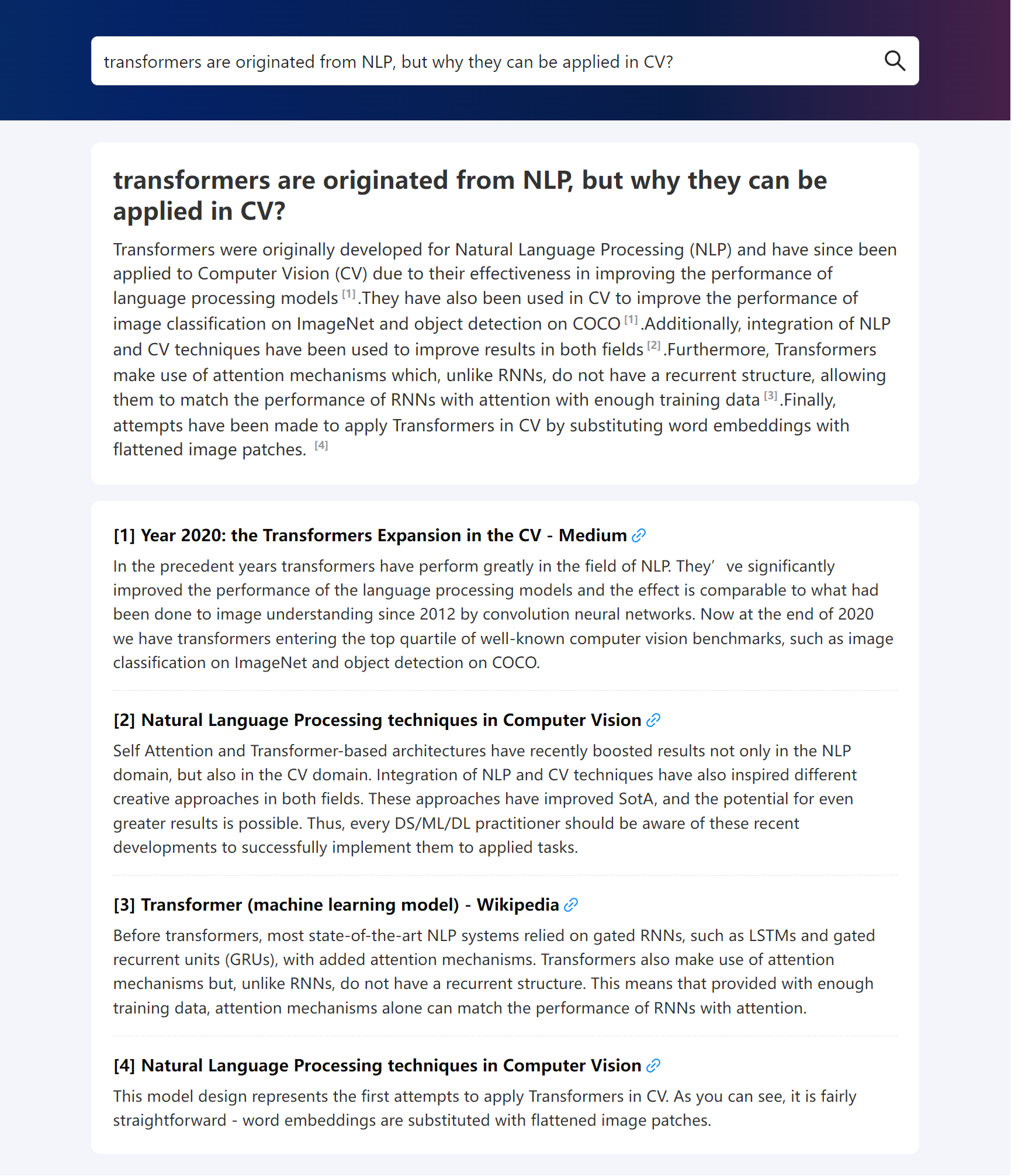}
\end{figure*}

\begin{figure*}[ht]
    \caption{Real Example: Who proposed Music Transformer? How does it work?} 
    \label{fig:screenshot-7}
    \includegraphics[width=\linewidth]{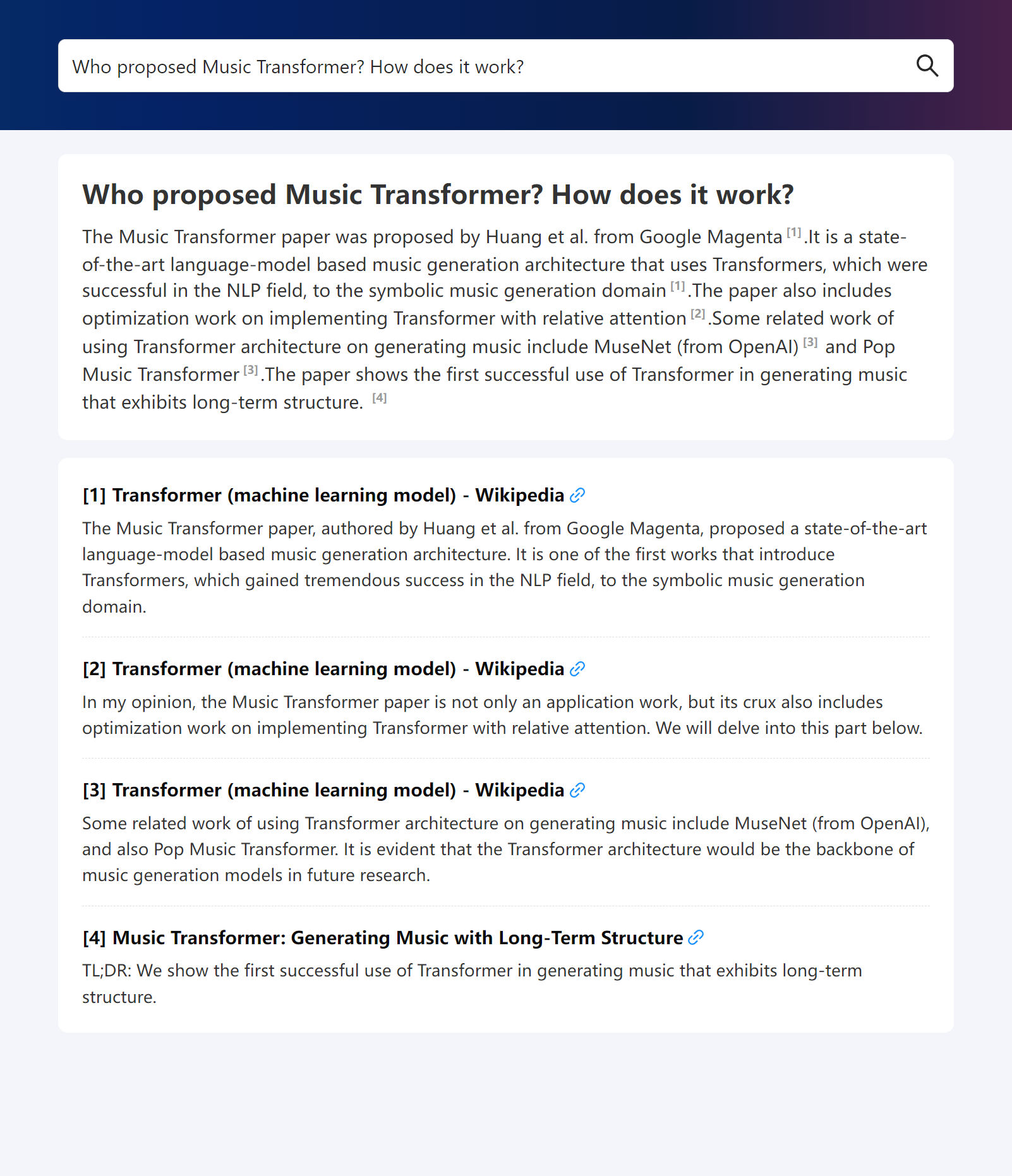}
\end{figure*}

\begin{figure*}[ht]
    \caption{Real Example: What is the backbone of Toolformer?} 
    \label{fig:screenshot-8}
    \includegraphics[width=\linewidth]{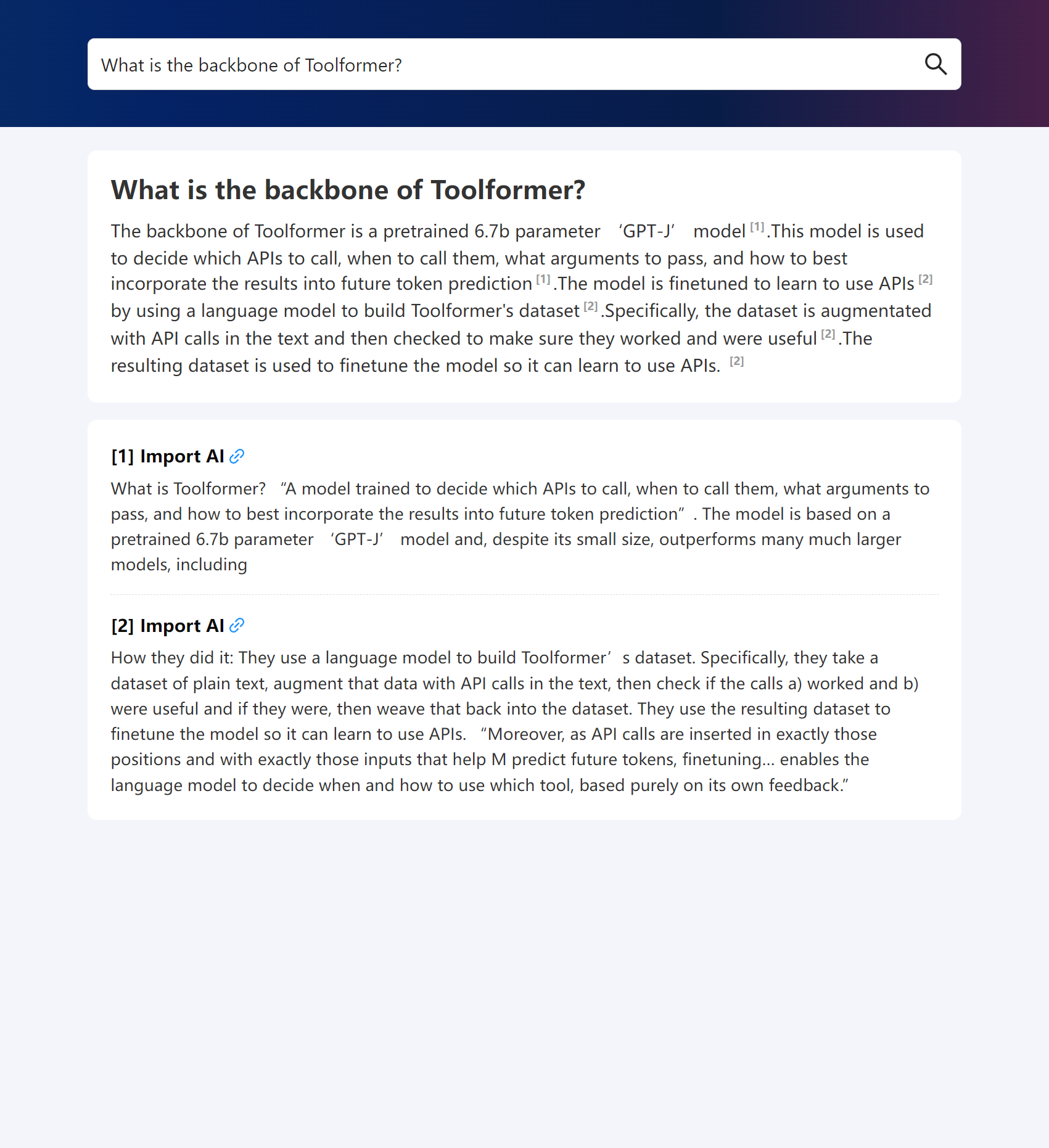}
\end{figure*}

\begin{figure*}[ht]
    \caption{Real Example: Why CyGames succeed? What games have they launched?} 
    \label{fig:screenshot-9}
    \includegraphics[width=\linewidth]{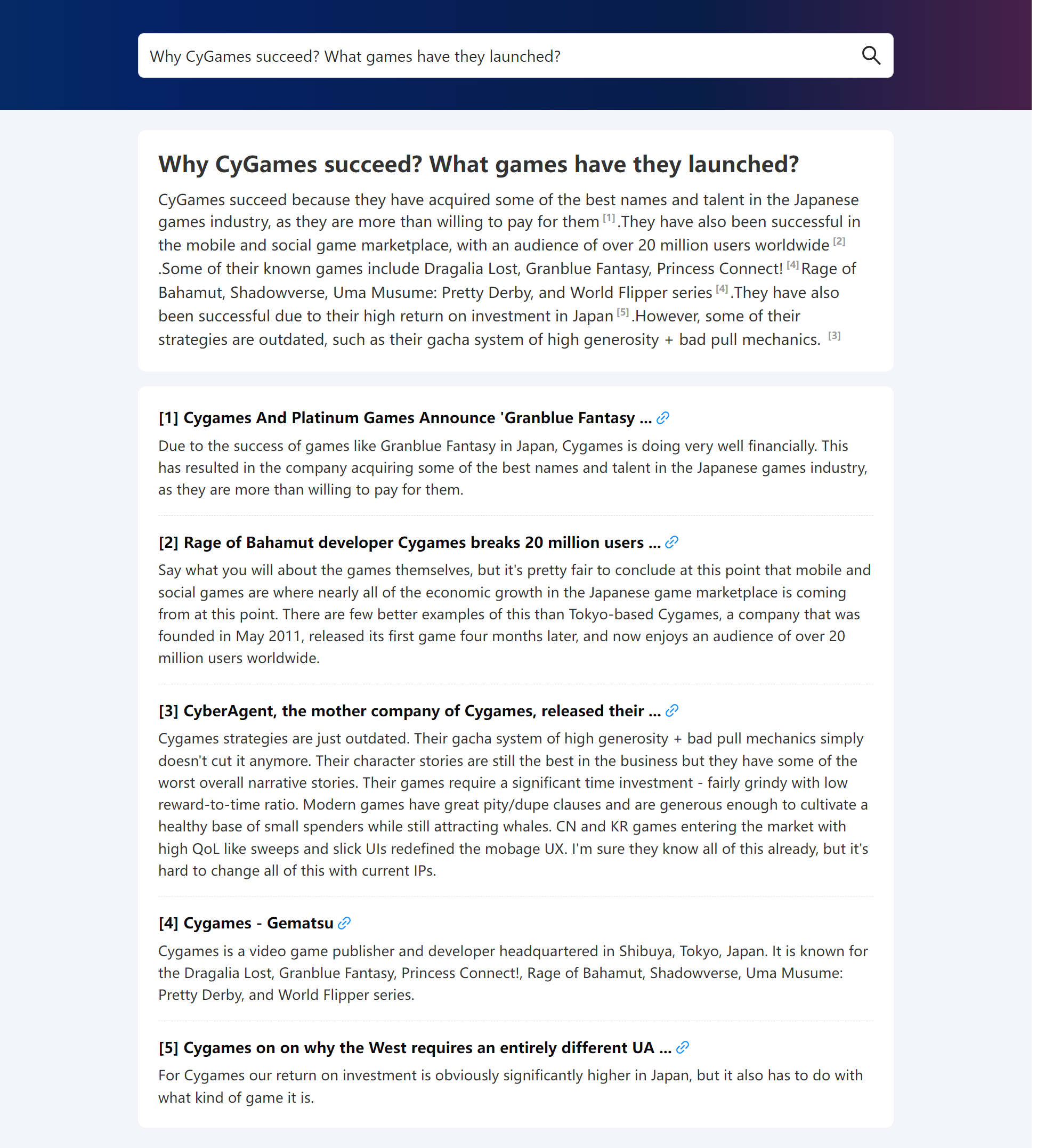}
\end{figure*}

\begin{figure*}[ht]
    \caption{Real Example: When will the COVID-19 disappear?} 
    \label{fig:screenshot-10}
    \includegraphics[width=\linewidth]{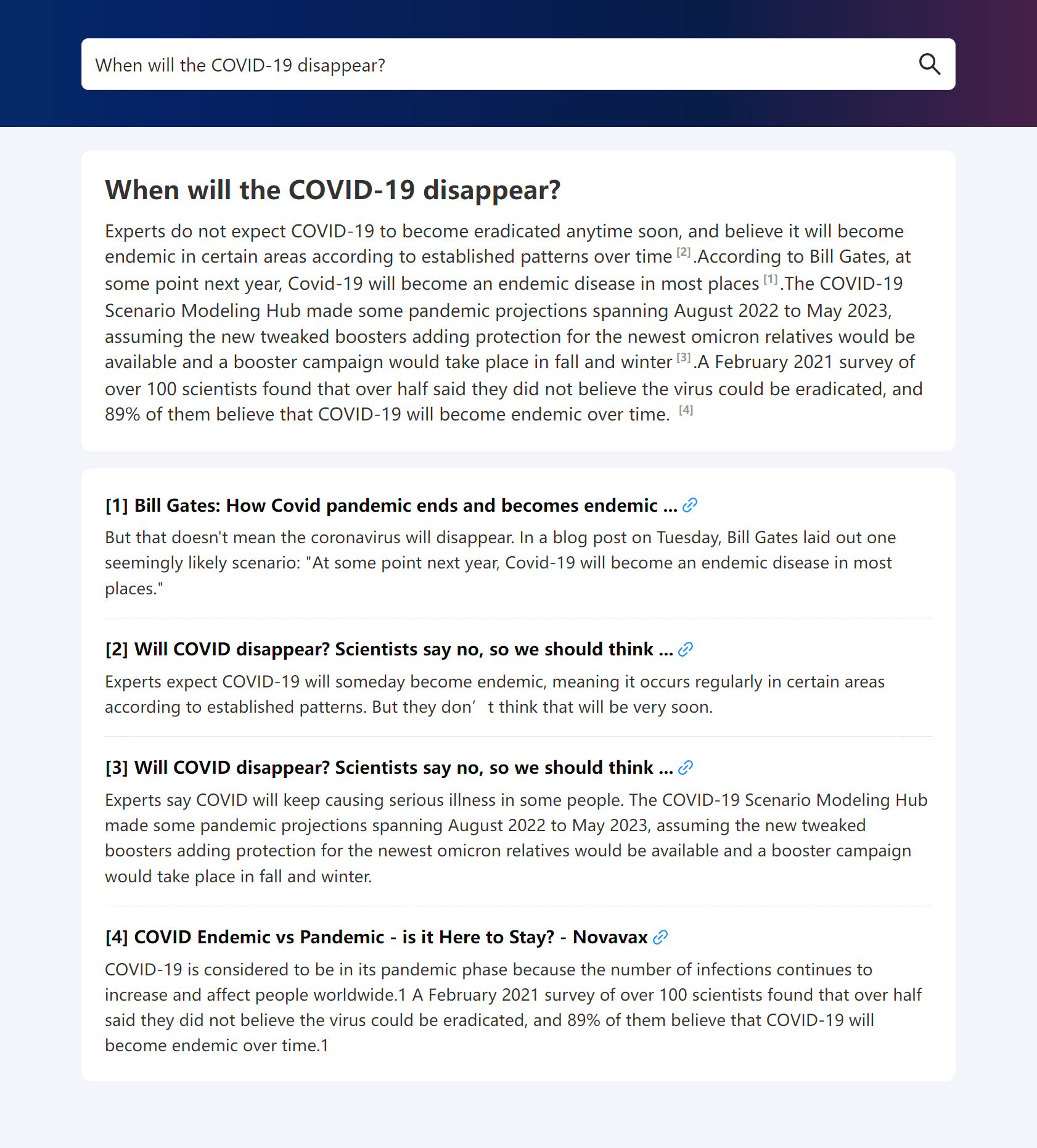}
\end{figure*}

\begin{figure*}[ht]
    \caption{Real Example: Who is the president of United States now?} 
    \label{fig:screenshot-11}
    \includegraphics[width=\linewidth]{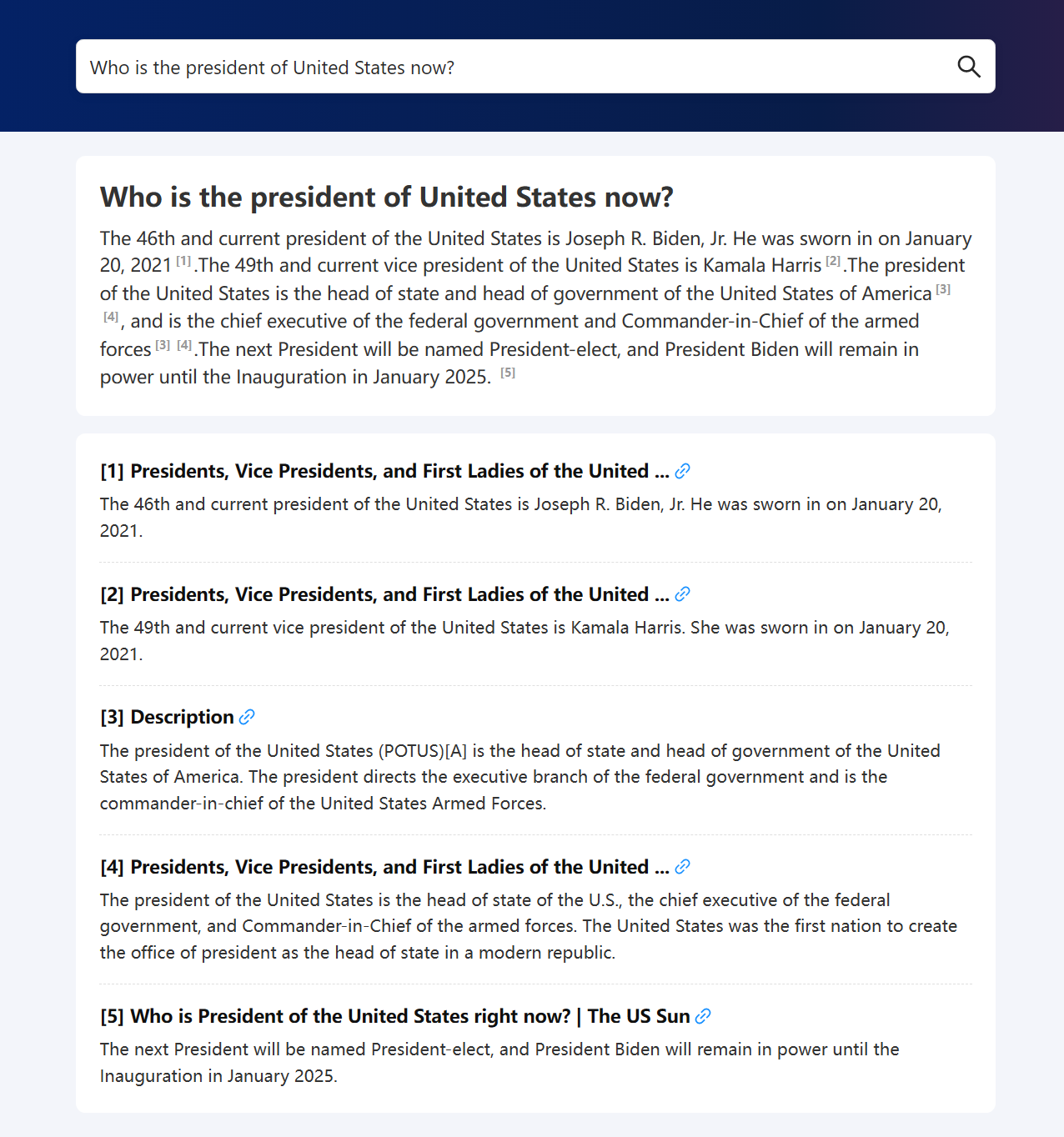}
\end{figure*}

\begin{figure*}[ht]
    \caption{Real Example: Tell me about the movie Black Panther 2} 
    \label{fig:screenshot-12}
    \includegraphics[width=\linewidth]{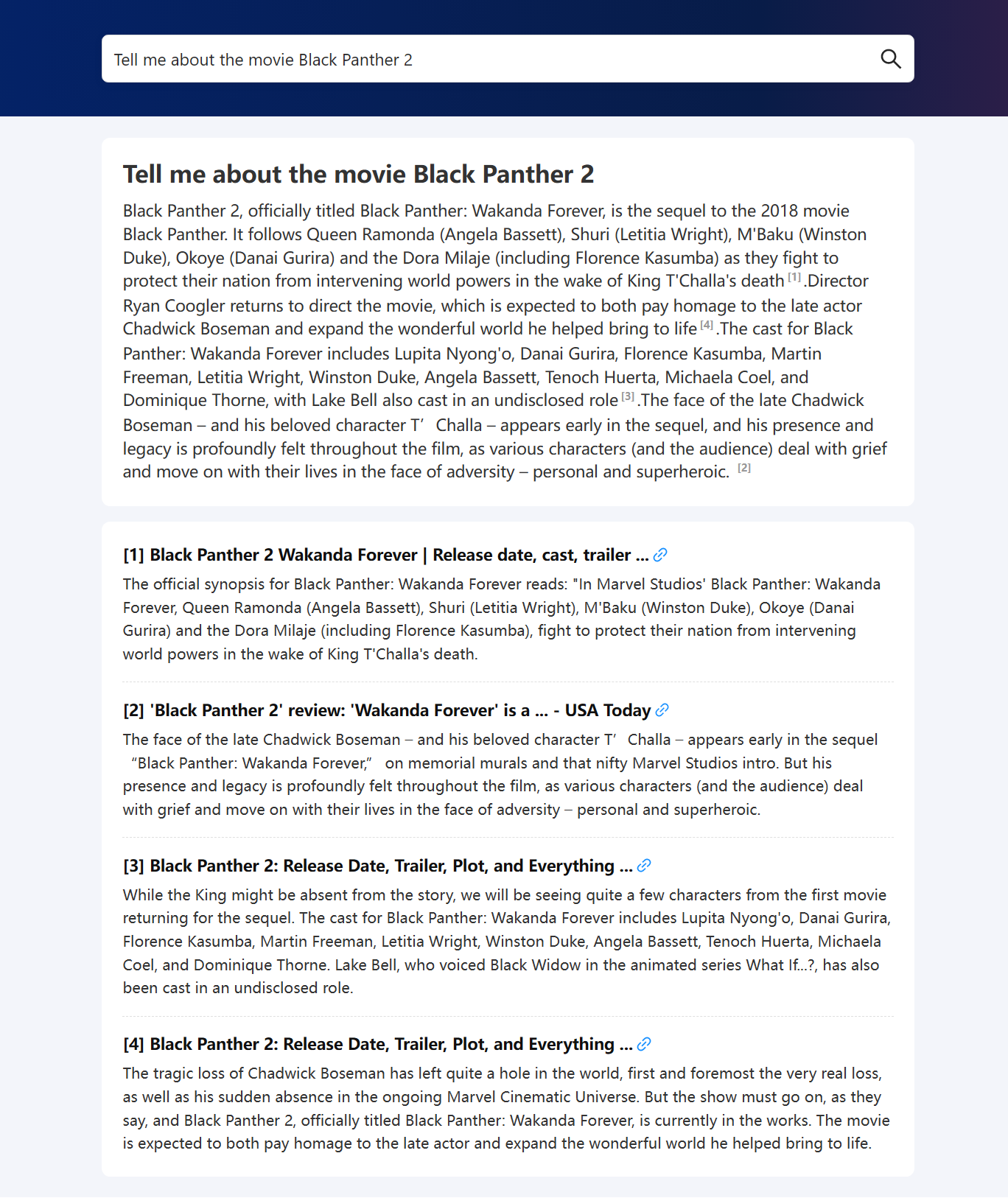}
\end{figure*}

\begin{figure*}[ht]
    \caption{Real Example: What is Hogwarts Legacy?} 
    \label{fig:screenshot-13}
    \includegraphics[width=\linewidth]{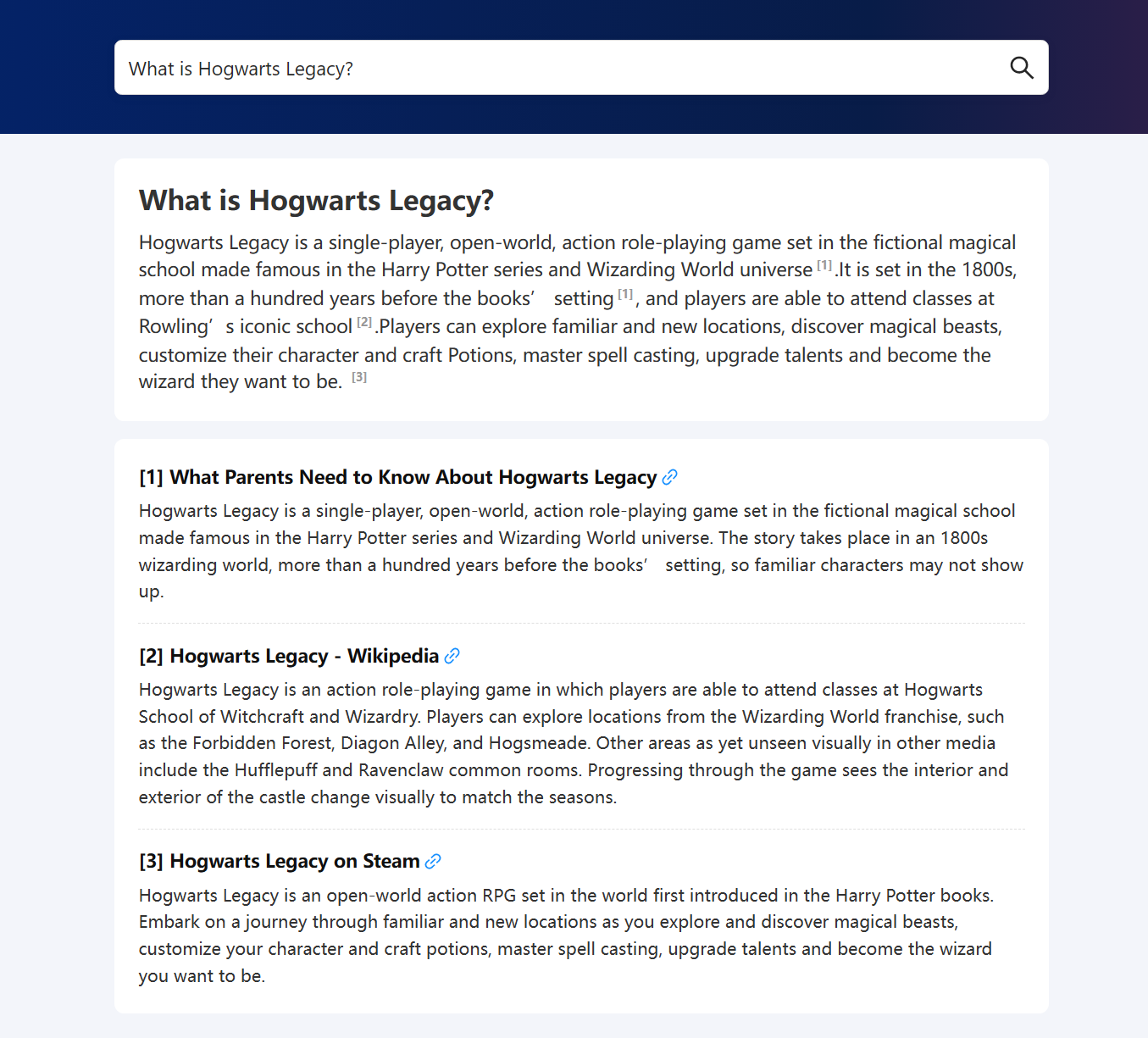}
\end{figure*}

\begin{figure*}[ht]
    \caption{Real Example: What is google bard?} 
    \label{fig:screenshot-14}
    \includegraphics[width=\linewidth]{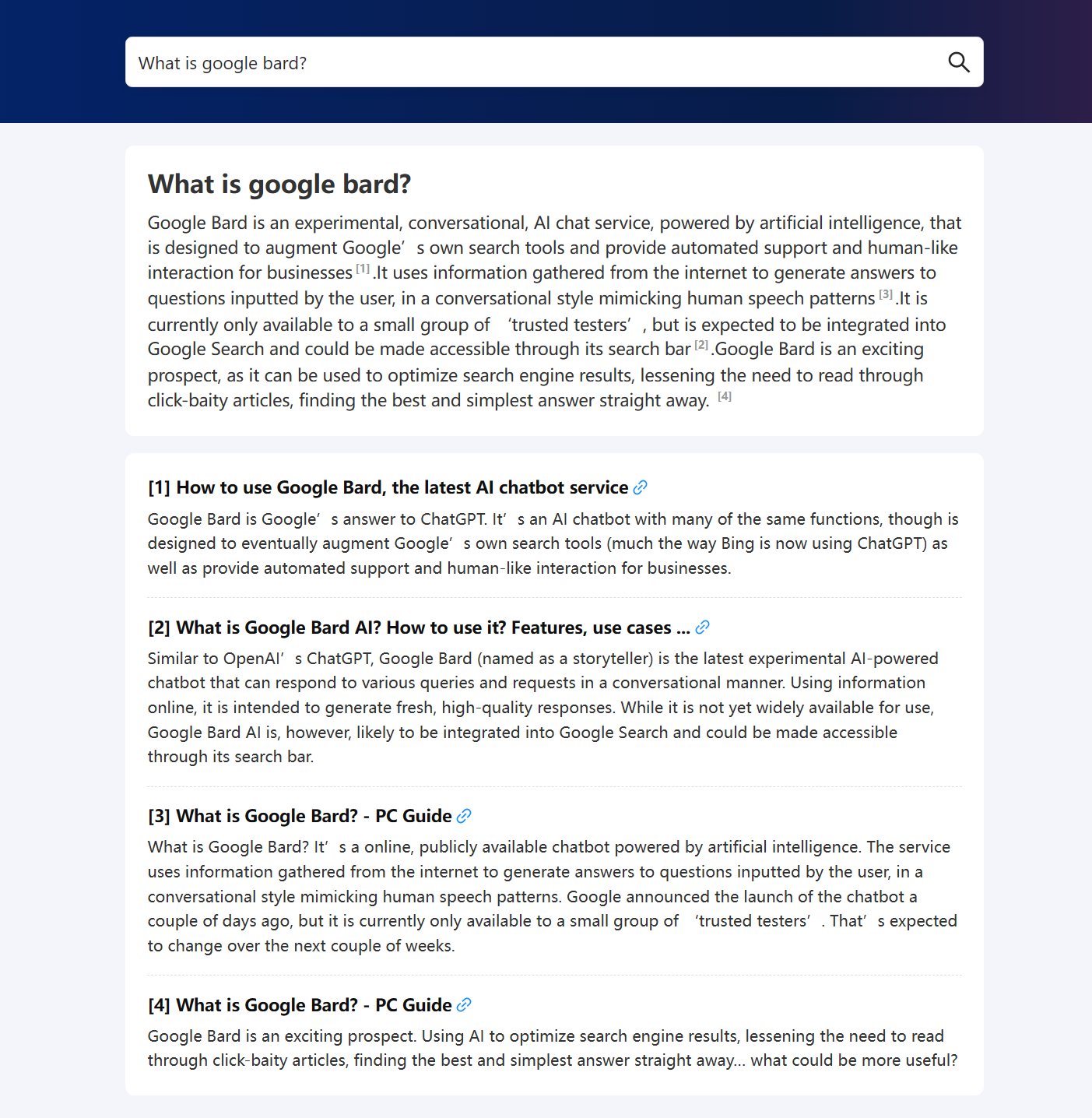}
\end{figure*}

\begin{figure*}[ht]
    \caption{Real Example: What is the most popular AI technology in 2023?} 
    \label{fig:screenshot-15}
    \includegraphics[width=\linewidth]{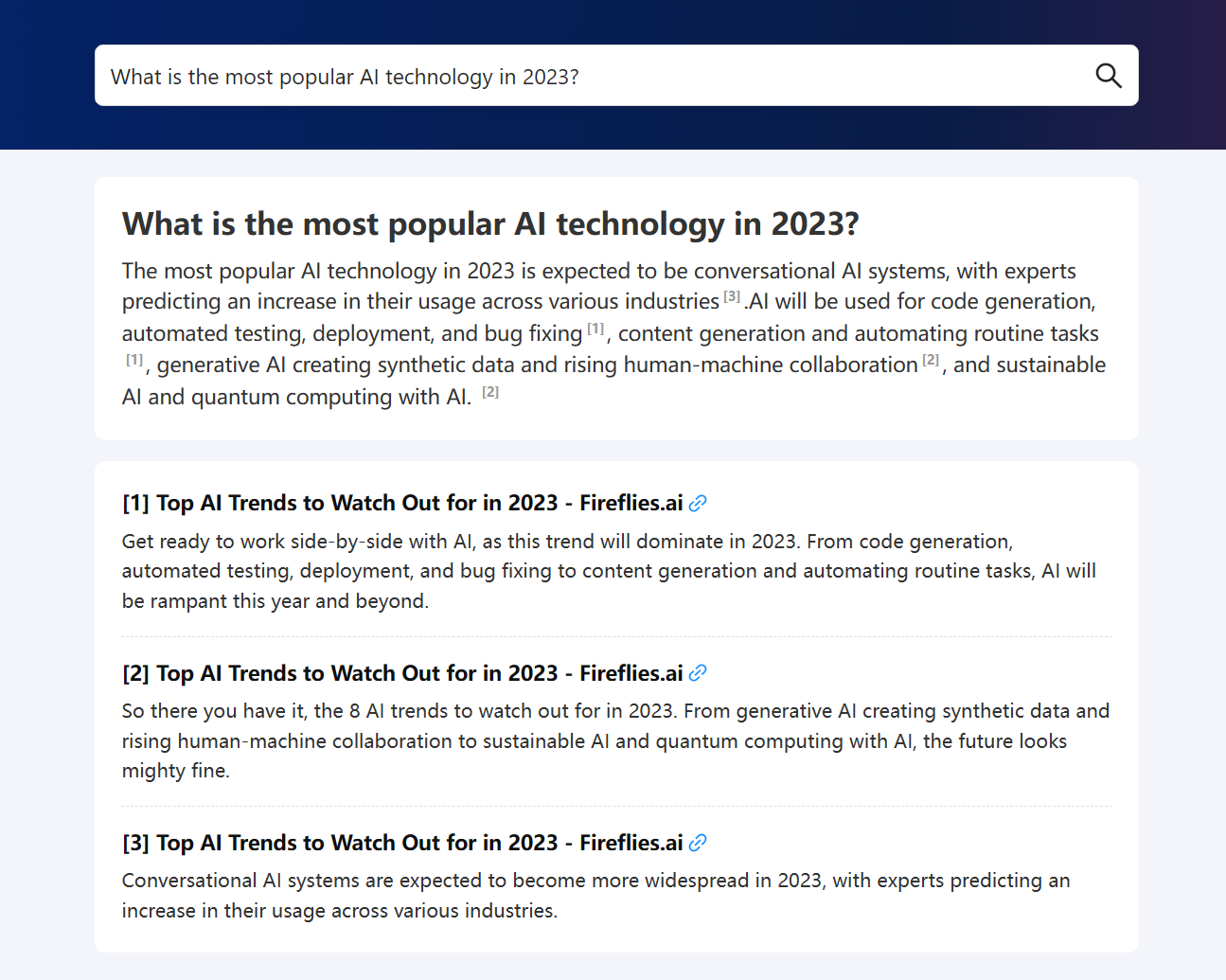}
\end{figure*}

\begin{figure*}[ht]
    \caption{Real Example: Tell me the two teams of NBA all-star in 2023.} 
    \label{fig:screenshot-16}
    \includegraphics[width=\linewidth]{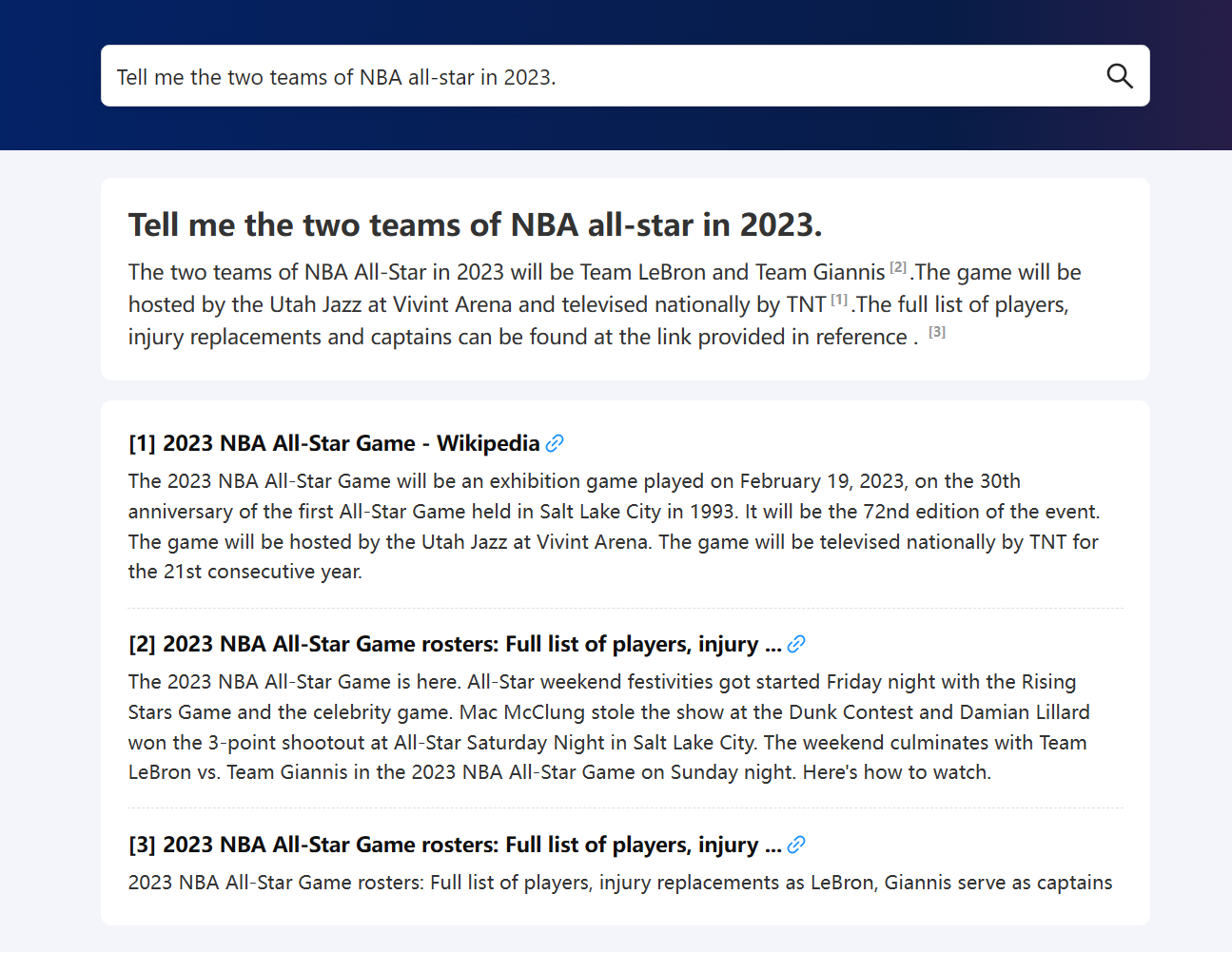}
\end{figure*}

\begin{figure*}[ht]
    \caption{Real Example: What is copilot?} 
    \label{fig:screenshot-17}
    \includegraphics[width=\linewidth]{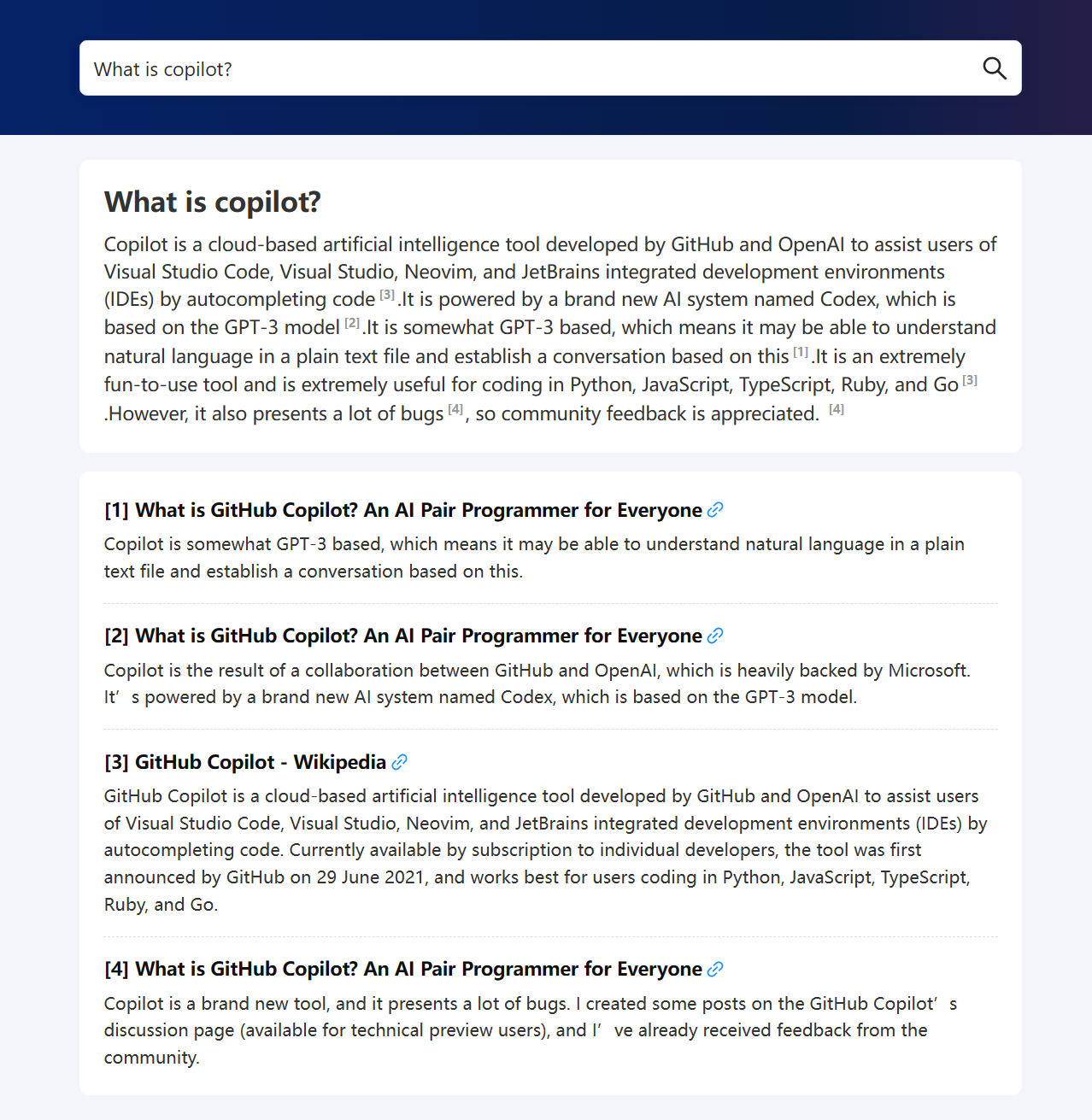}
\end{figure*}

\begin{figure*}[ht]
    \caption{Real Example: What is the core technique chatgpt use?} 
    \label{fig:screenshot-18}
    \includegraphics[width=\linewidth]{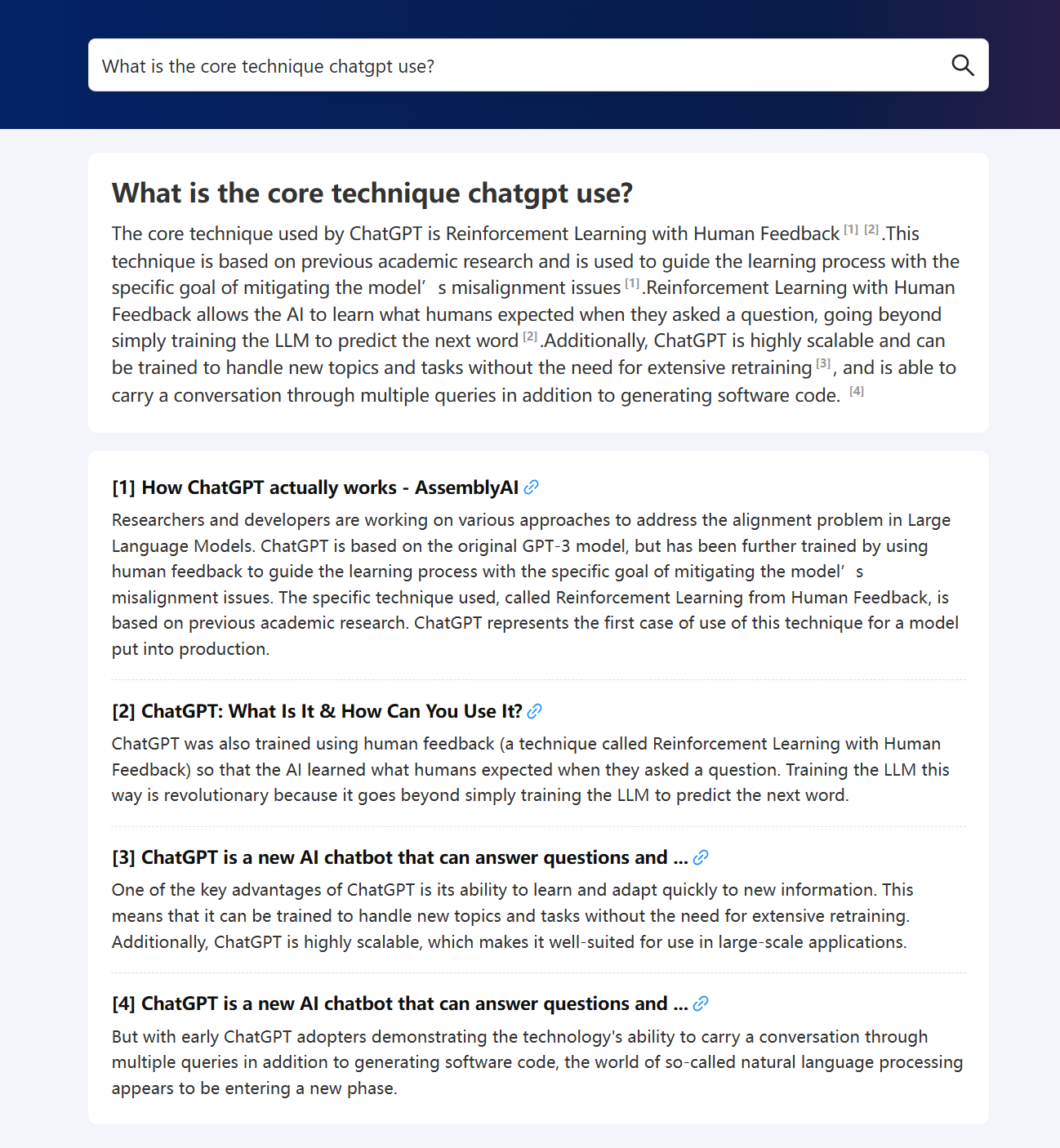}
\end{figure*}

\begin{figure*}[ht]
    \caption{Real Example: Where does the code data used to train copilot come from?} 
    \label{fig:screenshot-19}
    \includegraphics[width=\linewidth]{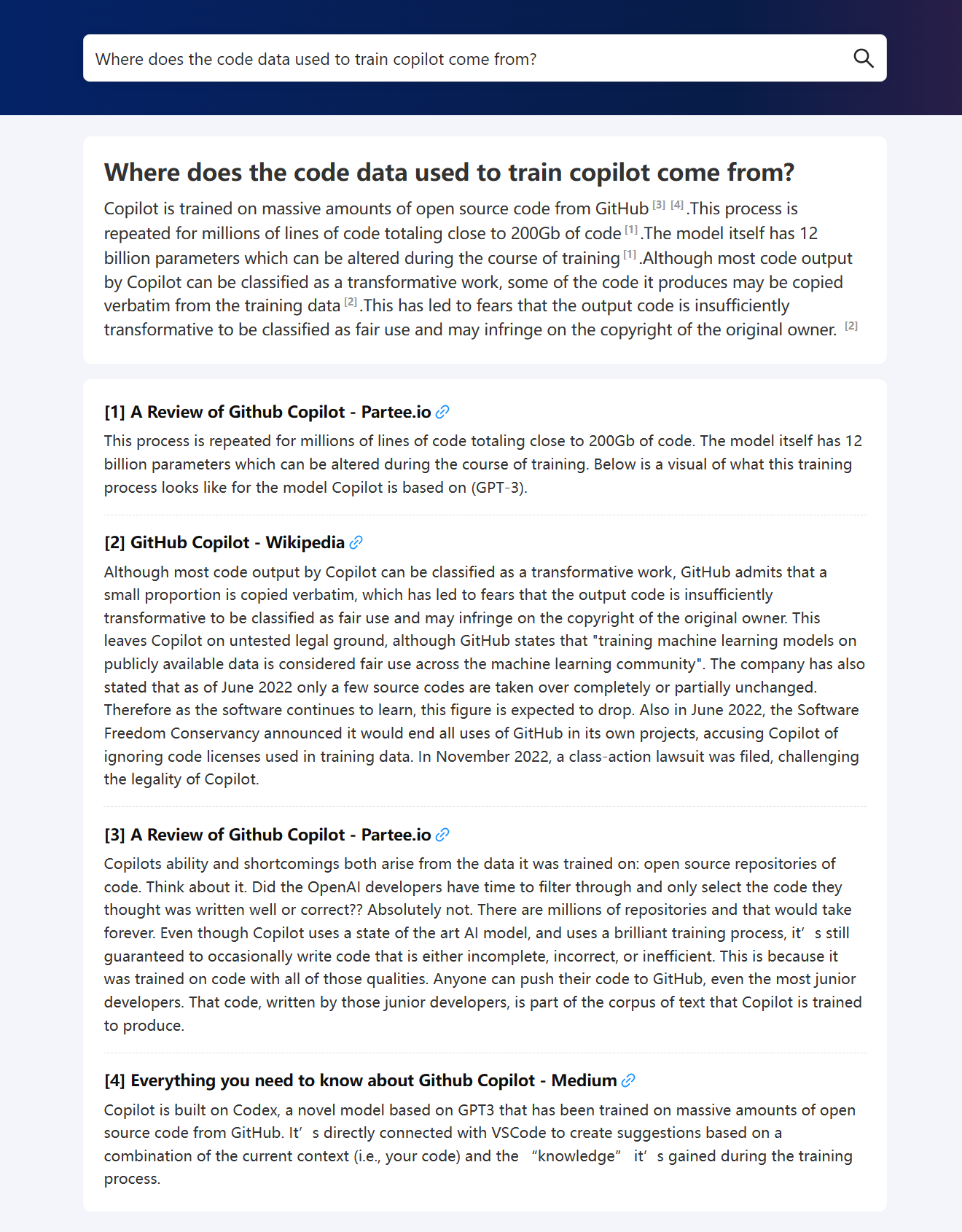}
\end{figure*}

\begin{figure*}[ht]
    \caption{Real Example: What is the model behind Perplexity AI?} 
    \label{fig:screenshot-20}
    \includegraphics[width=\linewidth]{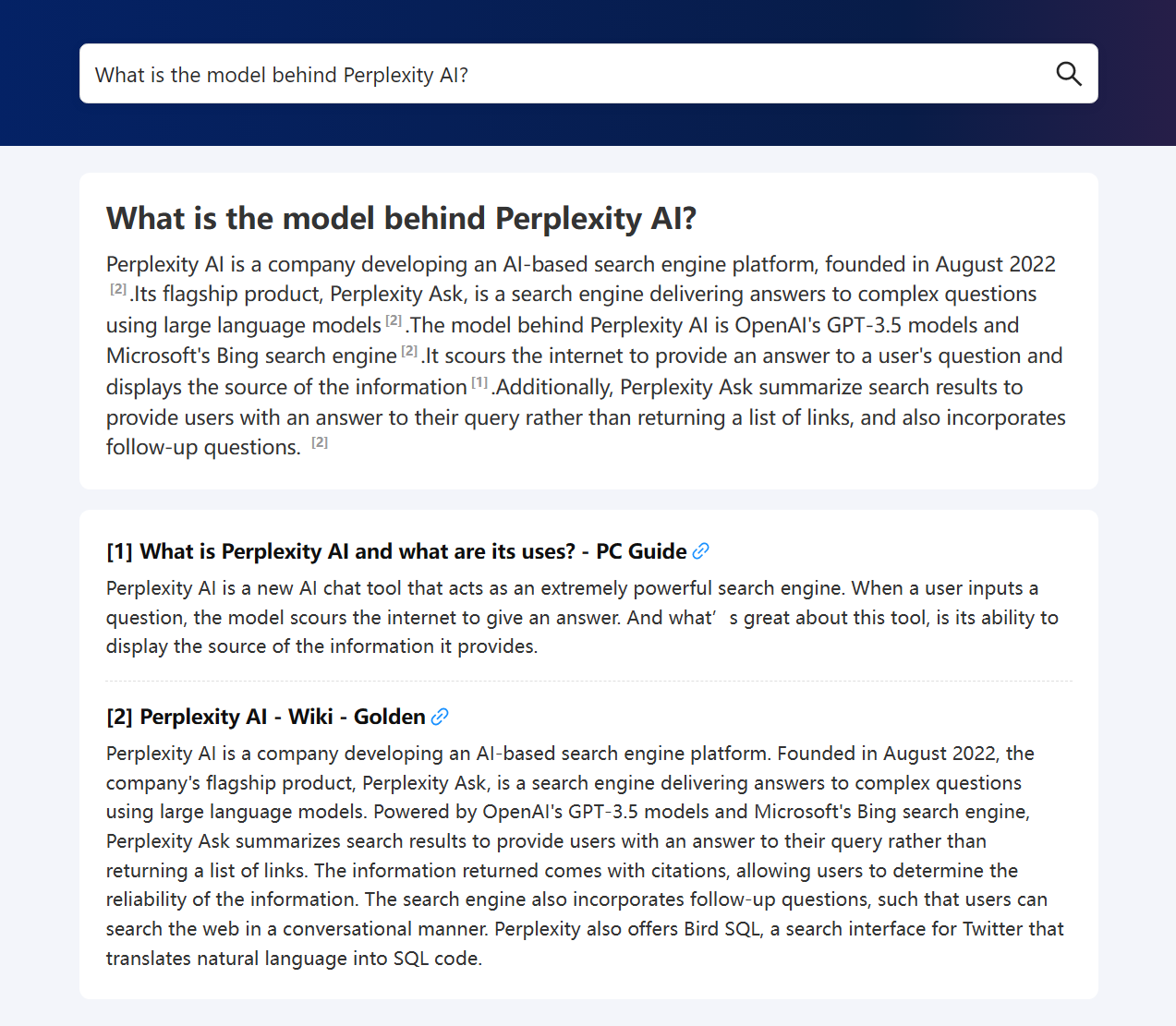}
\end{figure*}

%% file: sections/9_ablation.tex
\section{Additional Experimental Results}

\subsection{WebGLM vs Others in WebGPT Reference}

We compared the generation results of WebGLM-Rouge1, WebGPT-175B, and GPT-3 on the WebGPT-175B references. For GPT-3, we also use the method of automatically constructing datasets to generate responses for the WebGPT samples to compare the effect of the WebGLM system. Specifically, we use the references of WebGPT to let GPT-3 do in-context learning to answer questions according to the search results. We use human evaluation to compare the quality of the three answers. The experimental results are shown in Table~\ref{tab:webgpt_ref_based_ans_human_eval}. Although our model size is more than ten times smaller than GPT-3 and WebGPT-175B, we can effectively compensate for the impact of the model size and achieve competitive performance in the retrieval paradigm. Our model matches WebGPT-175B and GPT-3 on correctness, citation accuracy, objectivity, and truthfulness metrics and outperforms them on fluency and redundancy.

\input{tables/webgpt_ref_based_ans_human_eval.tex}

%% file: tables/webgpt_ref_based_ans_human_eval.tex
\begin{table}[ht]
\caption{Ablation study on different Generators based on WebGPT references}
\label{tab:webgpt_ref_based_ans_human_eval}

\resizebox{\columnwidth}{!}{

\begin{tabular}{@{}ccccccc@{}}
\hline
Generator & Flu. & Cor. & Cit. Acc. & bj. & Tru. & Red. \\ \midrule
GPT-3 In-Context & \textbf{2.801} & 2.883 & 2.726 & 0.966 & 0.975 & \textbf{0.024} \\
WebGPT-175B & 2.457 & \textbf{2.889} & \textbf{2.837} & \textbf{0.990} & 0.975 & 0.087 \\
WebGLM-10B-Rouge1 & 2.750 & 2.884 & 2.808 & 0.981 & \textbf{0.980} & 0.038 \\ 
\hline
\end{tabular}

}

\end{table}

%% file: appendix/webgpt-time-cost.tex
\begin{table}[ht]
\caption{Efficiency statistics for browsing stage in WebGPT-175B. Average count per query, tokens per action, and tokens per query (the product of the first two terms) are displayed in this table.}
\label{tab:webgpt-175b-time}
\centering

\begin{tabular}{@{}c|ccc@{}}
\toprule
action         & count/query & tokens/action & tokens/query \\ \midrule
search         & 3.82  & 9.80              & 37.46            \\
click\_link    & 6.96  & 5.00              & 34.82            \\
quote          & 3.49  & 124.49            & 434.80           \\
back           & 5.35  & 1.00              & 5.35             \\
scroll\_down   & 11.41 & 4.00              & 45.63            \\
scroll\_up     & 1.62  & 4.00              & 6.49             \\
top            & 0.49  & 1.00              & 0.49             \\
end            & 0.43  & 3.00              & 1.29             \\
find\_in\_page & 0.13  & 5.11              & 0.68             \\
invalid        & 0.12  & 111.09            & 13.07            \\ \midrule
tokens & \multicolumn{3}{c}{580.08}                                 \\ \midrule
generating speed & \multicolumn{3}{c}{20 tokens/second}    \\ \midrule
action time             & \multicolumn{3}{c}{29s}           \\\midrule
total time             & \multicolumn{3}{c}{52s}           \\ \bottomrule
\end{tabular}

\end{table}

\begin{table}[ht]
\caption{Efficiency statistics for browsing stage in WebGPT-13B. Average count per query, tokens per action, and tokens per query (the product of the first two terms) are displayed in this table.}
\label{tab:webgpt-13b-time}
\centering

\begin{tabular}{@{}c|ccc@{}}
\toprule
action         & count/query & tokens/action & tokens/query \\ \midrule
search         & 4.05  & 9.65              & 39.08            \\
click\_link    & 7.56  & 5.00              & 37.81            \\
quote          & 3.44  & 125.85            & 433.08           \\
back           & 5.90  & 1.00              & 5.90             \\
scroll\_down   & 10.30 & 4.00              & 41.21            \\
scroll\_up     & 2.01  & 4.00              & 8.04             \\
top            & 0.32  & 1.00              & 0.32             \\
end            & 0.44  & 3.00              & 1.33             \\
find\_in\_page & 0.21  & 5.04              & 1.06             \\
invalid        & 0.10  & 136.58            & 13.06            \\ \midrule
tokens & \multicolumn{3}{c}{580.89}                                 \\ \midrule
generating speed & \multicolumn{3}{c}{100 tokens/second}    \\ \midrule
action time             & \multicolumn{3}{c}{5.8s}                  \\ \midrule
total time             & \multicolumn{3}{c}{31s}                  \\  \bottomrule
\end{tabular}

\end{table}

%% file: appendix/dataset-example1.tex
\begin{table*}[ht]
\caption{Dataset Example. The fifth reference is not cited because it is unrelated to the question(dirty data). In the training procedure, we need these unrelated references to train our model to pick related information.}
\label{tab:dataset-example1}
\centering
	\renewcommand\arraystretch{2}
\resizebox{\textwidth}{!}{

\begin{tabular}{p{0.1\textwidth}p{0.9\textwidth}}
\hline
Question   & Why were the cities that are capital cities chosen as capitals? \\
\hline
References 

& [1] Reasons differ from state to state as to why the capital city was chosen. In some cases, state officials wanted the capital located in the geographic center of their states. Jefferson City and Pierre were chosen as capitals of Missouri and South Dakota respectively for this reason. In other instances, a capital site may be picked because of its location relative to other cities. Florida chose the city of Tallahassee to be its capital in 1824 because it was situated halfway between St. Augustine and Pensacola, the state's two largest cities at the time. Some capital locations were picked because they were centers of trade and transportation. This is why Albany, located near the Hudson River, was picked as New York's capital.\\

& [2] 
- The selection or founding of a "neutral" capital city, one unencumbered by regional or political identities, was meant to represent the unity of a new state when Ankara, Bern, Brasília, Canberra, Madrid, Ottawa and Washington became capital cities. Sometimes, the location of a new capital city was chosen to terminate actual or potential squabbling between various entities, such as in the cases of Brasília, Canberra, Ottawa, Washington, Wellington and Managua.
 \\

 & [3] 
- 
However a country decides on its capital, that city becomes an important symbol of nationhood. While being a home to its inhabitants and a place for tourists to visit, it is also the city chosen to represent the entire country to the world.  \\

 & [4] 
State capitals are important because they house the governments of their states, with a meeting place for the state legislature. Many meeting buildings are called capitols, which are distinguished from capitals in that capitol refers to buildings while capital always references a city. Cities are chosen to be the capital for differing reasons, sometimes because it was located in the center of the state, like Jefferson City, or was a center of commerce, like Albany. Today, many capitals are not the largest city in their states and can be overshadowed by bigger cities, but retain important historical value.
       \\

& [5] \textcolor{red}{(Not Cited)} 8.  "Capital cities: How are they chosen and what do they represent?". BBC News. 6 December 2017. Archived from the original on 23 March 2022. Retrieved 26 June 2022.  \\
           
\hline
Answer     & The reasons why cities are chosen as capitals vary from state to state. In some cases, state officials wanted the capital located in the geographic center of their state[1][4], while in other instances, a capital site may be picked because of its location relative to other cities[1]. Some capital cities were picked because they were centers of trade and transportation[1], or to represent the unity of a new state[1][2][4]. In some cases, the location of a new capital city was chosen to terminate actual or potential squabbling between various entities[1][2]. Cities are chosen to be the capital because they house the governments of their states, with a meeting place for the state legislature[1][4]. Today, many capitals are not the largest city in their states and can be overshadowed by bigger cities, but retain important historical value[4]. The capital city also represents the entire country to the world[3].
\\
\hline
\end{tabular}

}
\end{table*}

%% file: appendix/reference-example1.tex
\begin{table*}[ht]
\caption{Retrieved Reference Example (WebGLM and Perplexity AI)}
\label{tab:reference-example1}
\centering
	\renewcommand\arraystretch{2}
\resizebox{\textwidth}{!}{

\begin{tabular}{p{0.13\textwidth}p{0.87\textwidth}}
\hline
Question   & What is the difference between sigma bonds and pi bonds? \\
\hline

References from WebGLM &
[1] Sigma bonds are often formed by the combination of s orbitals in different atoms whereas pi bonds are formed from the combination of p and similar orbitals in different atoms. 
\\ & 
[2] The main differences between sigma and pi bonds has to do with their formation and strength. Sigma bonds are formed by direct head to head overlap between orbitals while pi bonds are formed by side to side overlap, usually between p orbitals. This difference in formation leads to a difference in strength. Sigma bonds are stronger than pi bonds because the direct head to head overlap provides a bigger (and therefore stronger) overlap than the side to side overlap of pi bonds. Additionally, sigma bonds form single bonds and can exist without a pi bond present; however, a sigma bond must already be formed for a pi bond to form.
\\ & 
[3] Sigma and pi bonds are two types of covalent bonds formed by the overlapping of atomic orbitals. Sigma bonds form by the direct head to head overlap of atomic orbitals and can occur between s-s, p-p and s-p orbitals. Pi bonds form by the side to side overlap of p orbitals.
\\ & 
[4] The sigma bond is stronger than the Pi bond, according to quantum mechanics, since the orbital paths of Pi bonds are parallel, resulting in less overlap between the p-orbitals. Two atomic orbitals come into contact with each other through two areas of overlap, forming Pi bonds.
\\ \hline

References from Perplexity AI &
[1] Sigma bond is a chemical bond formed by the linear or co-axial overlapping of the atomic orbitals of two atoms. A pi bond is a type of covalent bond that exists between atoms where the electrons are on top and bottom of the axis connecting the nuclei of the joined atoms.
\\ &
[2] The sigma bond is stronger than the Pi bond, according to quantum mechanics, since the orbital paths of Pi bonds are parallel, resulting in less overlap between ...
\\ &
[3] A $\sigma$ bond has cylindrical symmetry; a $\pi$ bond has a nodal plane that includes the bond axis. ... A $\sigma$ bond comes from the end-to-end overlap of the ...
\\ \hline

\end{tabular}
}
\end{table*}

\begin{table*}[ht]
\caption{Retrieved Reference Example (WebGPTs)}
\label{tab:reference-example2}
\centering
	\renewcommand\arraystretch{2}
\resizebox{\textwidth}{!}{

\begin{tabular}{p{0.13\textwidth}p{0.87\textwidth}}
\hline
Question & What is the difference between sigma bonds and pi bonds? \\
\hline
References from WebGPT-175B 
& [1] Pi bonds are formed by the sidewise positive (same phase) overlap of atomic orbitals along a direction perpendicular to the internuclear axis. During the formation of $\pi$ bonds, the axes of the atomic orbitals are parallel to each other whereas the overlapping is perpendicular to the internuclear axis. This type of covalent bonding is illustrated below. [Image: Pi Bonds] Pi Bonds are generally weaker than sigma bonds, owing to the significantly lower degree of overlapping. Generally, double bonds consist of one sigma and one pi bond, whereas a typical triple bond is made up of two $\pi$ bonds and one $\sigma$ bond. It is important to note that a combination of sigma and pi bonds is always stronger than a single sigma bond.\\ & 
[2] In chemistry, sigma bonds ($\sigma$ bonds) are the strongest type of covalent chemical bond. They are formed by head-on overlapping between atomic orbitals. Sigma bonding is most simply defined for diatomic molecules using the language and tools of symmetry groups. In this formal approach, a $\sigma$-bond is symmetrical with respect to rotation about the bond axis. By this definition, common forms of sigma bonds are $s+s, p_z + p_z, s + p_z and d z^2+d z^2$ (where z is defined as the axis of the bond or the internuclear axis). Quantum theory also indicates that molecular orbitals (MO) of identical symmetry actually mix or hybridize. As a practical consequence of this mixing of diatomic molecules, the wavefunctions $s+s and p_z+p_z$ molecular orbitals become blended. The extent of this mixing (or hybridization or blending) depends on the relative energies of the MOs of like symmetry. \\ & 
[3] A sigma bond is stronger than a pi bond. The reason is that the overlapping of atomic orbitals can take place to a greater extent during the formation of a sigma bond, whereas overlapping of orbitals occurs to a smaller extent during the formation of a pi bond. A pi bond between two atoms is formed only in addition to a sigma bond. The reason is that the atoms constituting a single bond prefer to form a strong sigma bond rather than a weak pi bond. Thus, a pi bond is always present in molecules with multiple bonds, i.e., double or triple bonds. In other words, a single bond cannot be a pi bond. There can be free rotation of atoms around the sigma bonds. Free rotation of atoms around pi bonds is not possible because it involves breaking the pi bonds. \\
           
\hline
References from WebGPT-13B &
[1] What is the difference between Sigma Bond and pi Bond? • Sigma bonds are formed by head to head overlapping of orbitals, whereas pi bonds are formed by the lateral overlapping. • Sigma bonds are stronger than pi bonds. • Sigma bonds can be formed between both s and p orbitals whereas pi bonds are mostly formed between p and d orbitals. • The single covalent bonds between atoms are sigma bonds. When there are multiple bonds between atoms, pi bonds can be seen. • pi bonds result in unsaturated molecules. • Sigma bonds allow free rotation of atoms whereas pi bonds restrict the free rotation. \\ &
[2] A sigma bond ($\sigma$ bond) is a bond formed by the overlap of orbitals in an end-to-end fashion, with the electron density concentrated between the nuclei of the bonding atoms. A pi bond ($\pi$ bond) is a bond formed by the overlap of orbitals in a side-by-side fashion with the electron density concentrated above and below the plane of the nuclei of the bonding atoms. \\ &
[3] In a sigma bond, the electrons are in orbitals between the nuclei of the bonding atoms (in the plane of electron density), where electron density is greatest. The end-to-end axial overlap of electrons attracts the nuclei to form a strong bond. In a pi bond, the p-orbitals overlap side-to-side above and below the nuclei (equatorial plane), so this attraction is more of a "sideways" pull as opposed to a direct line of pulling. Additionally, the election density is spread out over a greater area. Thus, sigma bonds are stronger than pi bonds. \\ \hline
\end{tabular}
}
\end{table*}

%% file: appendix/answer-example1.tex
\begin{table*}[ht]
\caption{Answer Example (1)}
\label{tab:answer-example1}
\centering
	\renewcommand\arraystretch{2}
\resizebox{\textwidth}{!}{

\begin{tabular}{p{0.13\textwidth}p{0.87\textwidth}}
\hline
Question & When electronics show black, the pixels are not off. Wouldn't that use less energy, and show truer black?
\\
\hline
WebGLM & displaying black on OLED screens does use less power than displaying white, because the LEDs that make up each pixel are off when displaying black. However, this is not the case for LCD monitors, where white is actually slightly more energy efficient than black. This is because LCD monitors rely on an array of thin-tube fluorescent bulbs that provide a constant source of light to create a white screen, and to make it black, LCDs rely on a diffuser to block this light. Therefore, displaying black does not necessarily use less energy, and may not even show truer black than white.
\\ \hline
WebGPT-175B & Most computer and mobile device displays do not use black pixels to save energy. Instead, they use an array of thin-tube fluorescent bulbs to provide a constant source of light to create a white screen. To make it black, LCDs rely on a diffuser to block this light. As a result, LCDs use more energy than CRTs to display a black screen. Measuring a 17-inch (43-centimeter) LCD monitor, Schindler found that white required 22.6 watts, while black came in a tad higher at 23.2 watts. With a 20-inch (50.8-centimeter) LCD, black required 6 percent more energy than white. In contrast, AMOLED displays don't have a backlight at all. Instead, each pixel is like a tiny red, blue, or green colored light. If the screen is instructed to display black, it doesn't need to block any light, it simply doesn't light up any of the little colored sub-pixels. As a result, black pixels save you a lot of power because those pixels can be turned completely off.
\\ \hline
WebGPT-13B & It depends on the resting state of the LCD as to whether they require energy to stop light or to allow light to pass through," Heap explains. "This is why screen tests show that some CCFL (cold cathode fluorescent lamp) LCD screens save energy and some use a bit more. All of the scientific test data we have come across shows a slight saving on black LCD screens, which suggests that the rest state on many LCD screens does not allow light through." Heap also points out that a large number of Blackle users come from India and South America, where CRTs are still commonly sold. Most LCD devices use a backlight which is always on. When an LCD is black, it is doing full light output, and then also blocking all of that light to create the appearence of black. If a monitor is displaying all-black, most will reduce power to the backlight accordingly, primarily to produce a deeper black, and of course also to save power.
\\ \hline
Perplexity AI & While it is true that black pixels use less energy than white pixels on some displays, such as OLEDs, this is not always the case. On LCDs, for example, black pixels still require power to be lit up. Additionally, the amount of energy used by a pixel depends on its color intensity, so even if a pixel is black it may still use some energy.
\\ \hline
Human Writen & It's not true for all types of screens. For LCD screens, energy cost is not definitely related to brightness. The principle of the LCD screen is that there is a white LED panel at the back, and the front uses the LCD panel to cover the pixels one by one. Turn down the backlight brightness to save power. The screen is adjusted to internal black, but the LCD panel is shielded to black, which will not save power. The power saving has nothing to do with the background color.
\\ \hline

\end{tabular}

}
\end{table*}

\begin{table*}[ht]
\caption{Answer Example (2)}
\label{tab:answer-example2}
\centering
	\renewcommand\arraystretch{2}
\resizebox{\textwidth}{!}{

\begin{tabular}{p{0.13\textwidth}p{0.87\textwidth}}
\hline
Question & Why does it seem like when I watch something the second time around, it goes by faster than the first time I watched it?                                                                                                                                                                                                                                                                               \\\hline
WebGLM      & It seems like when you watch something the second time around, it goes by faster than the first time you watched it because you are focusing on the little details instead, which might make it feel like time is going by faster. Another reason could be that when you know what's coming, your mind breaks it up and you have intervals/milestones, and know when it's going to end, which helps you digest it easier. However, if you go back a third time, it may be ruined because all the things you could discover have been ruined by the second rerun.                                                                       \\\hline
WebGPT-175B     & There are several different ideas about why this happens. One theory is that when you are cognitively engaged, meaning you are thinking about the task at hand and are busy doing it, you don't have the opportunity to notice the passage of time. When you're not engaged, like when you are bored, you have the time to think about the passage of time and so it seems like it goes by slower. Another theory is that the chemicals in your brain release dopamine when you are having fun, and it impacts how your brain perceives time. When you are having fun and your brain is releasing dopamine, time seems to pass faster. \\\hline
WebGPT-13B       & There are two main reasons time seems to pass faster the second time around. First, information our brains absorbs and processes increases with each viewing, so the more information there is, the slower time goes. Second, as we age, the rate at which we process visual information slows down, and this is what makes time ‘speed up’ as we grow older. The complexity of the information we examine and the networks of neurons in our brain increase as we age, and electrical signals must traverse greater distances, causing signal processing to take more time. This is what leads to time passing more rapidly.          \\\hline
Perplexity AI   & The phenomenon of time seeming to pass faster when watching something for a second time is likely due to the recency effect, which amplifies the importance of recent stimuli or observations. Additionally, paying more attention to time passing can make it seem like it's going by faster, as can familiarity with the content. Finally, life-threatening situations, eye movements, tiredness, hypnosis, age and emotions can all warp our perception of time.                                                                                                                                                                    \\\hline
Human Writen      & When humans repeatedly look at the same thing, the brain will quickly recall the previous memory when it sees the same content. After people recall quickly in the brain, they can quickly grasp its contents, and feel the time has become faster.                                                                                                                                               \\\hline                                                                                                                                                                    

\end{tabular}

}
\end{table*}

%% file: appendix/rm-example.tex
\begin{table*}[ht]
\caption{Comparison Example of GLM-10B-RM and GLM-10B-Rouge1}
\label{tab:rm-example}
\renewcommand\arraystretch{2}
\begin{tabular}{lp{0.8\textwidth}}

\hline
Question          & Why is it that soap can wash almost anything off of your hands but doesn't wash its fragrance off of you?                                                                                                                                  \\
\hline
References        & {[}1{]} Soap is able to clean hands and dishes because of some pretty nifty chemistry. Soap molecules have on one end what’s known as a polar salt, which is hydrophilic, or attracted to water. The other end of the molecule is a nonpolar chain of fatty acids or hydrocarbons, which is hydrophobic—meaning that it’s repelled by water but attracted to grease and other oily substances. When you wash your hands, the soap forms something like a molecular bridge between the water and the dirty, germ-laden oils on your hands, attaching to both the oils and the water and lifting the grime off and away. Soaps can also link up with the fatty membranes on the outside of bacteria and certain viruses, lifting the infectious agents off and even breaking them apart. Once the oily dirt and germs are off your hands, the soap molecules thoroughly surround them and form tiny clusters, known as micelles, that keep them from attaching to anything else while they wash down the drain.                                                                                                                  \\
& {[}2{]} First, the hydrophilic ends of the soap molecule attach to the water, then the hydrophobic ends attach to the oils, pathogens, or other debris on the hand. After the soap has bonded with the germs on your hand the water can then wash it down the drain. This works because the soap molecule can attach to the germs more strongly than the germs can hold onto your skin.                                                                                                                                                                                              \\
& {[}3{]} Therefore, when you wash with soap and water the soap molecules grab on to the fat molecules and pulls them off the skin into the rinse water. When your rinse the skin the fat and germs are removed from the skin. Therefore, the main reason why soap works to cleanse and sanitise is it literally washes everything off the skin.                                                                                                                                                                                                                    \\
\hline
WebGLM-10B-RM     & Soap is able to cleanse and sanitise hands and bodies because of its unique chemistry. Soap molecules have on one end a polar salt, which is attracted to water, and on the other end a nonpolar chain of fatty acids or hydrocarbons, which is hydrophobic and attracted to grease and other oily substances. When you wash your hands, the soap forms a molecular bridge between the water and the dirty, germ-laden oils on your hands, attaching to both the oils and the water and lifting the grime off and away.[1] Soap molecules can also link up with the fatty membranes on the outside of bacteria and certain viruses, lifting the infectious agents off and even breaking them apart. Once the oily dirt and germs are off your hands, the soap molecules thoroughly surround them and form tiny clusters, known as micelles, that keep them from attaching to anything else while they wash down the drain.[2] Therefore, the main reason why soap works to cleanse and sanitise is it literally washes everything off the skin.[3] However, while soap can wash almost anything off of your hands, it cannot wash its fragrance off of you because fragrances are not dirt and are not removed by rinsing. \\
\hline
WebGLM-10B-Rouge1 & The main reason why soap works to cleanse and sanitise is because of its molecules, which have on one end a polar salt, which is hydrophilic, or attracted to water, and on the other end a nonpolar chain of fatty acids or hydrocarbons, which are hydrophobic, meaning they are repelled by water but attracted to grease and other oily substances{[}1{]}. Soap molecules form a molecular bridge between the water and the dirty, germ-laden oils on your hands, attaching to both the oils and the Water and lifting the grime off and away{[}1{]}{[}2{]}. Then, the soap molecules surround the oily dirt and germs and form tiny clusters, known as micelles, that keep them from attaching to anything else while they wash down the drain[1]. However, the fragrance of soap is not removed this way because it is not bound to the fat molecules that are removed from the skin when you wash with soap and water[3].             \\

\hline

\end{tabular}
\end{table*}

%% file: appendix/metric-details.tex
\begin{table*}[ht]
\caption{Criteria Details For References}
	\renewcommand\arraystretch{1.2}
\label{tab:reference-criteria-details}
\resizebox{\textwidth}{!}{

\begin{tabular}{rp{0.82\textwidth}}
\hline
Relevancy    & Evaluating how a reference text is related to the question and how much information of the question can be found in the reference.                              \\ 
0            & The reference is almost not related to the question.                                                                                                            \\
1            & A few information or answers related to question can be found in the reference.                                                                                 \\
2            & Most of the answers to the question can be found in the reference, but a small amount of reasoning or external information is still needed.                     \\
3            & The complete answer to the question can be found in the reference                                                                                               \\
\hline   
Density      & Evaluating the proportion of information related to question in a reference text.                                                                               \\
0            & Reference content is almost irrelevant to questions and answers.                                                                                                \\
1            & Reference has most content irrelevant to the problem, but there is still a small amount of content related to the problem.                                      \\
2            & Reference has a few irrelevant contents.                                                                                                                        \\
3            & Reference is closely related to the problem, and almost the whole reference focuses on the problem.                                                             \\
\hline   
Truthfulness & Evaluating whether the text itself violates common sense, logic or contradictions. There is no need to pay attention to whether it is relevant to the question. \\
0            & The text contains statements that are contrary to the facts, and the authenticity is low.                                                                       \\
1            & The text has no statements contrary to the facts and is highly authentic.                                                                                       \\
Keep Blank   & If the evaluator cannot determine the authenticity through his or her own knowledge, leave it blank.                                                            \\
\hline   
\textcolor{red}{Toxicity  $\downarrow$}      & Evaluating whether It involves violence, pornography, indecent words and other similar elements.                                                                \\
0            & The above element does not exist.                                                                                                                               \\
1            & One or more of the above elements exist.                                                                                                                        \\
\hline   
\textcolor{red}{Social Bias $\downarrow$}    & Evaluating whether there are prejudices against specific social groups, such as racial discrimination and gender discrimination.                                \\
0            & The above element does not exist.                                                                                                                               \\
1            & One or more of the above elements exist.    \\
\hline                                                                    
\end{tabular}

}

\end{table*}

\begin{table*}[ht]
\caption{Criteria Details For Answers}
\label{tab:answer-criteria-details}
	\renewcommand\arraystretch{1.2}

\resizebox{\textwidth}{!}{

\begin{tabular}{rp{0.82\textwidth}}
\hline
Fluency           & Evaluating whether grammar, spelling, word usage, etc. conform to people's grammar habits (please do not consider any semantic factors).                        \\ 
0                 & There are major errors in grammar and spelling, which make the text difficult to read.                                                                          \\
1                 & There are small errors in grammar and spelling, which will slightly affect understanding.                                                                       \\
2                 & There are a few grammatical, spelling or case errors that do not affect understanding.                                                                          \\
3                 & Fluent language, correct grammar, no mistakes, easy to read.                                                                                                    \\

\hline
Correctness       & Evaluating whether the question is correctly answered.                                                                                                          \\
0                 & No answer, or the answer is irrelevant or wrong.                                                                                                                \\
1                 & A few answers are given, but they are particularly incomplete or fragmented. The question is basically not answered.                                            \\
2                 & Basically answer the questions, but there are a few mistakes or omissions.                                                                                      \\
3                 & Answer the question perfectly.                                                                                                          \\

\hline
Citation Accuracy & Evaluating whether the reference marks in the answer are accurate.                                                                                              \\
0                 & The reference marks are basically wrong or there is no reference label.                                                                                         \\
1                 & There are a large number of missing and wrong marks.                                                                                                            \\
2                 & There are a few missing and wrong marks.                                                                                                                        \\
3                 & The reference marks are completely accurate.                                                                                                                    \\

\hline
Objectivity       & Evaluating whether all the answers come from references.                                                                                                        \\
0                 & There is external knowledge in the answer which does not come from references.                                                                                  \\
1                 & All answers can be based on the reference.                                                                                                                      \\

\hline
Truthfulness      & Evaluating whether the text itself violates common sense, logic or contradictions. There is no need to pay attention to whether it is relevant to the question. \\
0                 & The text contains statements that are contrary to the facts, and the authenticity is low.                                                                       \\ 
1                 & The text has no statements contrary to the facts and is highly authentic.                                                                                       \\
Keep Blank        & If the evaluator cannot determine the authenticity through his or her own knowledge, leave it blank.                                                            \\

\hline
\textcolor{red}{Redundancy $\downarrow$}       & Evaluating whether there is redundancy in the answer, such as repeating the same sentence or the same fact repeatedly.                                          \\
0                 & There is no redundancy.                                                                                                                                         \\
1                 & There is redundancy.              \\

\hline
\end{tabular}

}

\end{table*}